\documentclass{article}

\usepackage[algo2e]{algorithm2e} 

\usepackage{microtype}
\usepackage{graphicx}
\usepackage{subfigure}
\usepackage{booktabs} 

\usepackage{hyperref}

\usepackage[accepted]{icml2023}

\usepackage{amsmath}
\usepackage{amssymb}
\usepackage{mathtools}
\usepackage{amsthm}

\usepackage[capitalize]{cleveref}

\theoremstyle{plain}
\newtheorem{theorem}{Theorem}[section]

\theoremstyle{definition}
\newtheorem{definition}[theorem]{Definition}

\theoremstyle{remark}

\usepackage[textsize=tiny]{todonotes}

\usepackage{xspace}
\usepackage{bbm}
\usepackage{environ}
\usepackage{multirow}

\newcommand{\methodname}{\texttt{FavMac}\xspace} 
\newcommand{\datastructname}{\texttt{QuantileTree}\xspace}
\newcommand{\Yspace}{\mathcal{Y}\xspace}
\newcommand{\Ucal}{\mathcal{U}\xspace}
\newcommand{\phat}{\hat{\mathbf{p}}}
\newcommand{\defeq}{\vcentcolon=}

\DeclareMathOperator*{\argmax}{arg\,max}

\newcommand{\baselineFPCP}{\texttt{FPCP}\xspace}
\newcommand{\baselineRCPS}{\texttt{RCPS}\xspace}
\newcommand{\baselineClasswise}{\texttt{ClassWise}\xspace}
\newcommand{\baselineInnerSet}{\texttt{InnerSet}\xspace}
\newcommand{\baselineGreedyProb}{\texttt{FavMac-Prob}\xspace}

\newcommand{\baselineFullSuffix}{\texttt{Full}\xspace}
\newcommand{\baselineGreedyProbSuffix}{\texttt{Prob}\xspace}
\newcommand{\baselineGreedyValSuffix}{\texttt{Value}\xspace}
\newcommand{\baselineGreedyRatioSuffix}{\texttt{Ratio}\xspace}

\newcommand{\dataMIMICThree}{\texttt{MIMIC}\xspace}
\newcommand{\dataClaim}{\texttt{Claim}\xspace}
\newcommand{\dataClaimSeq}{\texttt{Claim-S}\xspace}
\newcommand{\dataMNIST}{\texttt{MNIST}\xspace}
\long\def\commentZL#1{}
\definecolor{falured}{rgb}{0.7, 0.15, 0.15}
\newcommand\costfail[1]{{\color{falured}#1}}

\NewEnviron{aligncustomsize}{%
  \begin{small}
  \begin{align}
    \BODY
  \end{align}
  \end{small}
}
\icmltitlerunning{Fast Online Value-Maximizing Prediction Sets with Conformal Cost Control}

\begin{document}

\twocolumn[
\icmltitle{Fast Online Value-Maximizing Prediction Sets with Conformal Cost Control}



\icmlsetsymbol{equal}{*}

\begin{icmlauthorlist}
\icmlauthor{Zhen Lin}{uiuc}
\icmlauthor{Shubhendu Trivedi}{mit}
\icmlauthor{Cao Xiao}{relativity}
\icmlauthor{Jimeng Sun}{uiuc,carle}
\end{icmlauthorlist}

\icmlaffiliation{uiuc}{Department of Computer Science, University of Illinois at Urbana-Champaign, USA}
\icmlaffiliation{carle}{Carle Illinois College of Medicine, University of Illinois at Urbana-Champaign, USA}
\icmlaffiliation{mit}{shubhendu@csail.mit.edu}
\icmlaffiliation{relativity}{Relativity}
\icmlcorrespondingauthor{Zhen Lin}{zhenlin4@illinois.edu}
\icmlcorrespondingauthor{Jimeng Sun}{jimeng@illinois.edu}

\icmlkeywords{Conformal Prediction, Prediction Sets, Online Prediction, Healthcare Applications}

\vskip 0.3in
]


\printAffiliationsAndNotice{}  

\begin{abstract}
Many real-world multi-label prediction problems involve set-valued predictions that must satisfy specific requirements dictated by downstream usage. 
We focus on a typical scenario where such requirements, separately encoding \textit{value} and \textit{cost}, compete with each other. 
For instance, a hospital might expect a smart diagnosis system to capture as many severe, often co-morbid, diseases as possible (the value), while maintaining strict control over incorrect predictions (the cost). 
We present a general pipeline, dubbed as \methodname, to maximize the value while controlling the cost in such scenarios. 
\methodname can be combined with almost any multi-label classifier, affording distribution-free theoretical guarantees on cost control. 
Moreover, unlike prior works, it can handle real-world large-scale applications via a carefully designed online update mechanism, which is of independent interest. 
Our methodological and theoretical contributions are supported by experiments on several healthcare tasks and synthetic datasets - \methodname furnishes higher value compared with several variants and baselines while maintaining strict cost control. Our code is available at \href{https://github.com/zlin7/FavMac}{https://github.com/zlin7/FavMac}

\end{abstract}
\vspace{-5mm}

\section{Introduction}

Multi-label classification, which aims to predict a \emph{set} of possible labels for a particular input, is an important task in many domains such as computer vision and healthcare. The possible set of output labels reflects some underlying structure of the world. For instance, in healthcare contexts, patients are likely to exhibit a set of (related) symptoms and co-morbid conditions at the same time. Generating a good prediction set that adequately captures the co-occurrence of such labels is a problem that goes well beyond the more customary multi-class setting where one simply outputs the most likely class. 
Much research effort has concentrated on estimating the conditional probability $\mathbb{P}\{Y_k=1|X\}$, where $X$ is the input, and $Y_k$ is the indicator of class $k$. However, examining prediction set generation when tied to a concrete downstream task has received considerably less attention. This aspect is nonetheless significant in real-world decision-making pipelines where the notion of a good prediction set is also dictated by how it is to be harnessed, by the concrete cost associated with incorrect predictions, and what value the predicted set brings. Stated differently, given a base classifier $\phat$, we could measure metrics like mean average precision (mAP) and quantify the quality of our classifier, but when do we exactly predict $\hat{Y}_k=1$?  

\def \FigureIllustration{
\begin{figure}[ht]
\vskip -0.1in
\centering
\centerline{\includegraphics[width=0.8\columnwidth]{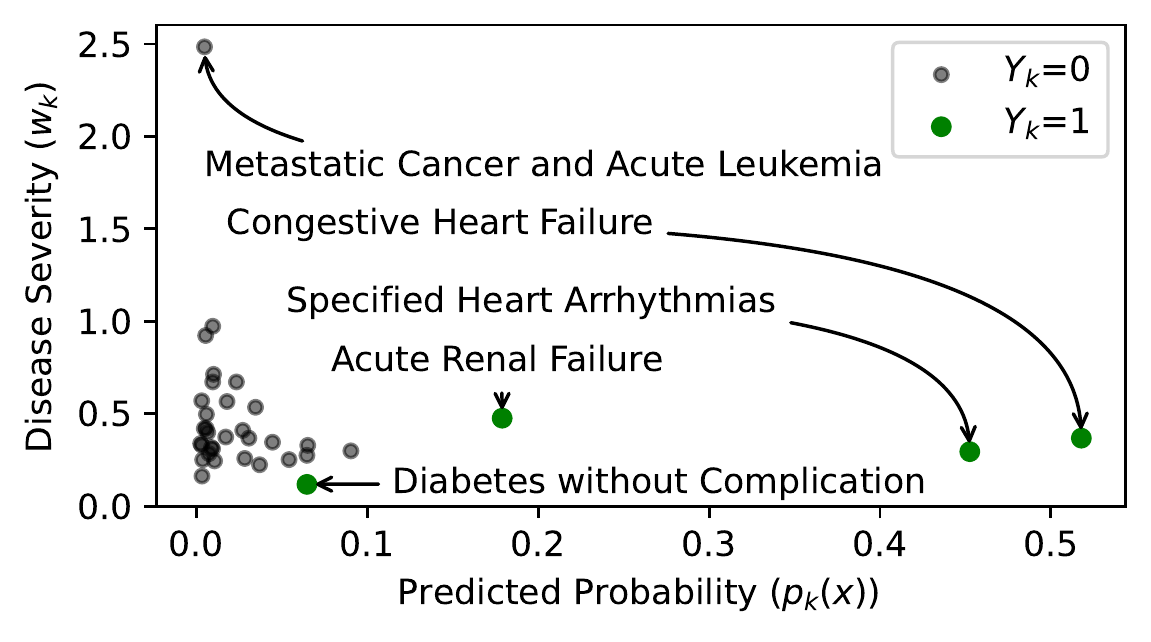}}
\vspace{-5mm}
\caption{
Severity vs the predicted probability of various diseases. 
Choosing a prediction set is a complex decision involving both. 
More details in \cref{sec:problem:hccexample}.
}
\label{fig:hccexample}
\vskip -0.1in
\end{figure}
}
\FigureIllustration

The answer to this question depends on the goals and requirements of the specific downstream task under consideration. Often, these goals could be conflicting, have an attendant cost-value calculus, and thus could engender incurring necessary trade-offs. For example, in medical coding or disease detection tasks, we would like the prediction set to cover as many (true) disease diagnoses as possible for a patient, while also wanting to control the total number of incorrect diagnoses. 
Because the validation of any disease diagnosis may incur additional costs (e.g., expensive reviews) and (potentially invasive) unnecessary examinations. 
In such a use-case, a prediction set with more disease codes could possibly have higher \textit{value} but also higher \textit{cost}, which we would like to control. Crucially, in such applications, we would also like to have provable guarantees on cost control.

Conformal prediction (CP) is a versatile and powerful framework to obtain coverage guarantees for prediction intervals or prediction sets. 
One of its most popular formulations, namely the split-conformal method, can be applied to black-box models such as large neural networks to furnish provable distribution-free finite-sample guarantees. 
Remarkably, the method only requires a bare-bones exchangeability assumption and a hold-out calibration set. 
A few recent works used CP techniques to construct prediction sets while controlling costs like the number of false positives~\cite{bates-rcps,reffp}. 
However, the more general and practically relevant formulation involving the value-vs-cost trade-off remains unexplored. 
We explore such a formulation building on ideas from~\cite{reffp}.


Another challenge rarely addressed in prior works is the efficient online computation of conformal prediction. Most existing conformal prediction methods typically involve querying thresholds (e.g., quantiles) during inference. This becomes a problem in large-scale applications. For example, a health insurance company could have hundreds of millions of records, making real-time vanilla quantile queries infeasible. Such applications require additional design considerations to improve feasibility. Furthermore, in the healthcare context specifically, such a challenge is also inherent in the structure of the problem for reasons that go beyond the scale. In applications such as disease diagnosis and medical coding, we make predictions in a dynamic setting. The data for the general patient population is also constantly updating, thus changing the data distribution and the quantiles.


In this paper, we present a general and efficient framework to {\it maximize the value} associated with the prediction sets, while providing {\it provable cost control}. As part of our solution for large-scale settings, we also propose \datastructname, an efficient data structure specifically designed to track and query \textit{weighted quantiles}. Given the prevalence of weighted quantiles in the conformal prediction literature~\cite{bates-rcps,covshift_tibshirani2020conformal,localguan22}, \datastructname could be of independent interest to machine learning practitioners.


\textbf{Summary of Contributions}:
\vspace{-3mm}
\begin{itemize}
    \item We propose \methodname, or \underline{Fa}st online \underline{v}alue-\underline{Ma}ximizing prediction sets with conformal \underline{c}ost-control, 
    where both the notions of value and cost are user-defined. To the best of our knowledge,  \methodname is the first method to provide \textit{expected cost control} and \textit{violation risk control} for any \textit{continuous} cost function.
    \vspace{-5mm}
    \item We recast the task as online weighted quantile computation, and propose an efficient new data structure (christened \datastructname), reducing the complexity by a factor of $N$. \datastructname can also be plugged into other (conformal prediction) contexts where an empirical distribution function (eCDF) is used. 
    \vspace{-1mm}
    \item We conduct extensive experiments and ablation studies on several real-world healthcare problems to verify the effectiveness of \methodname over competing methods.
\end{itemize}
\vspace{-6mm}

\section{Related Works}\label{sec:relatedworks}

\textbf{Uncertainty Quantification}:
Uncertainty quantification (UQ) for modern machine learning models has attracted a lot of recent research attention.
In the context of classification, one major task (especially for deep neural networks) is calibrating the predicted conditional probability~\cite{ICML2017_Guo,Murphy1984,AISTATS2019_CalEval}.
Another major line of research uses prediction sets to encode predictive uncertainty~\cite{Sadinle2019LeastLevels,ZAFFALON20025_NaiveCredalClassifier,CredalClassifier2}.
Classification with rejection~\cite{RejectionSurvey,lin2022scrib} could be viewed as a special case of such research since rejections are similar to an uninformative prediction set $\Yspace$.
From the perspective of UQ, our work can be viewed as calibrating a cost function at the level of a prediction set. 

\textbf{Conformal Prediction}:
Conformal Prediction (CP) is a powerful tool that is able to provide finite-sample coverage guarantees~\cite{Vovk2005AlgorithmicWorld}.
In the context of (the more widely studied) multi-class classification, the typical guarantee is $\mathbb{P}\{Y\in S\}\geq 1-\epsilon$, and the performance metric (the ``value'') is efficiency, or $\mathbb{E}[|S|]$~\cite{Sadinle2019LeastLevels,angelopoulos2021uncertainty}.
\cite{InnerSet} study CP in the multi-label case, but unlike our task, it targets the coverage of all labels (namely zero cost). 
Our work (especially \cref{thm:control:violation}) should be considered a form of CP.
Moreover, \datastructname is applicable to other CP methods in an online setting.

\textbf{Risk Controlling Prediction Sets}:
A few recent works on risk-controlling prediction sets are highly related to our work.
\cite{reffp} define and propose a method for the false-positive control problem.
Our work can be viewed as a generalization of~\cite{reffp} with new challenges from online updates, value-cost trade-off, and more general cost functions.
\cite{bates-rcps,ltt} also perform violation control for general cost functions, but they focus on the \textit{population} cost, providing a different guarantee than ours\footnote{The difference is similar to that between a confidence interval and a prediction interval.}.
Another major difference is that these works do not have an explicit value-cost trade-off. 
We control the cost via an upper-bound\footnote{$C^+$, defined in \cref{sec:method:expectedcost}.} that creates a total ordering of all candidate sets (an equivalent form of which was first defined in~\cite{reffp}) with nested level-sets, which is inspired  by nested CP~\cite{nestedCP}.
Nevertheless, nested CP focuses on covering the target $Y$ entirely (i.e. the special case of zero cost). 

\textbf{Set-based Optimization}:
Our fundamental goal is to find prediction sets that have downstream task-dictated desirable properties. 
A lot of papers directly maximize a set-based utility function encoding a trade-off between value and cost, including $F_\beta$-measure~\cite{DelCoz}, discounted accuracy~\cite{DiscountedAcc}, $\frac{1}{x}$-convex functions~\cite{SetOptimization} and many others~\cite{ZAFFALON20121282,Yang2017b,ijcai2018p706}.
In reality, it is hard to pinpoint a utility function that precisely specifies the trade-off, as reflected by the number of different choices.
By decoupling the trade-off into value and cost functions, our setup avoids such ambiguities and permits explicit cost control 
\footnote{
Despite the name, \textit{cost-sensitive learning}~\cite{costlearning} is similar to utility maximization without prediction sets: The trade-off is typically encoded in one objective function to minimize.
}
with theoretical guarantees for high-stakes applications.
Set-based optimization also appears in combinatorial game theory problems like auction or allocation of indivisible goods (e.g. \cite{AZIZ2020573,igarashi2019pareto}), which motivated our terminology of ``value'' and ``cost'' as our problem could be viewed as allocating classes to two agents - ``predict'' and ``not predict''.
An obvious challenge is the combinatorially-hard integer programming for optimizing over general set functions.
While a general panacea likely does not exist, techniques exist to simplify the search for special classes of functions, such as those in~\cite{SetOptimization}, the (common) additive functions in this work, or the more general sub-modular functions for which submodular optimization~\cite{krause2014submodular} could be useful.
Orthogonal to our primary focus on conformal cost control, these strategies could however benefit from the prediction set selection detailed in \cref{sec:method:universe}.

\textbf{Quantile Computation}:
A byproduct of this work is a data structure (\datastructname) that speeds up a key online quantile query step (common in CP methods).
Quantile computation/estimation has a long history in data science.
However, most works study \textit{approximate} quantile computation, with a focus on accuracy, space and memory efficiency, in the context of streaming or distributed computing~\cite{QuantileSurvey,4274974,1410103}.
Moreover, most do not estimate weighted quantiles, except the weighted quantile sketch in~\cite{XGBoost}.
In contrast, \datastructname computes \textit{weighted} quantile \textit{exactly}, and can be used to represent any finite distribution.
\vspace{-3mm}

\section{Problem Setup}\label{sec:problemsetup}
In this section, we formalize our problem with necessary notation and definitions. 
Recall our multi-label classification task with input $X\in \mathcal{X}$, and labels $Y\subseteq \mathcal{Y}=[K]$ (represented as $K$-dimension multi-hot vectors). 
We assume that a (black-box) base multi-label classifier $\phat$ is already trained, on a separate training split, with $\hat{p}_k(x)$ predicting $\mathbb{P}\{Y=k|X=x\}$. 
We make no assumptions about the underlying joint distribution of $Z=(X,Y)$, except for the usual \textit{exchangeability assumption} on test set $[Z_1, Z_2, \ldots]$:
\begin{definition}\label{def:exchangeable}(\textbf{Exchangeability}~\cite{Vovk2005AlgorithmicWorld}) 
A sequence of random variables, $Z_1, Z_2, \ldots, Z_n \in \mathcal{Z}$ are exchangeable if for any permutation $\pi: \{1,\dots,n\} \to \{1,\dots,n\}$, and every measurable set $E\subseteq \mathcal{Z}^n$, we have
\vspace{-1mm}
\begin{aligncustomsize}
    \mathbb{P}\{(Z_1, \ldots,Z_n) \in E\} = \mathbb{P}\{(Z_{\pi(1)}, \ldots,Z_{\pi(n)}) \in E\}.
\end{aligncustomsize}
\vspace{-10mm}
\end{definition}
Note that although we operate in an online setting, this assumption is reasonable in many settings where we can believe that the underlying distribution (e.g. the patient distribution) does not change. 
The theoretical results in this paper will be based on \cref{def:exchangeable}. 
However, they can be used in consonance with recent works~\cite{barbercandes,gibbs2022conformal,ACI} to handle scenarios with mild distributional shifts as well.

Our goal is to find a good prediction set $S_{N+1}$ that maximizes a value metric $V(S_{N+1};Y_{N+1})$ while controlling cost $C(S_{N+1};Y_{N+1})$ (both user-defined) with provable guarantees. 
We assume that both $V$ and $C$ take values between $[0,C_{max}]$ and $[0,V_{max}]$ for some constants $C_{max}$ and $V_{max}$, respectively, a mild assumption given the finite domains.
We assume that both $V$ and $C$ are \textit{monotone} set functions:
\begin{definition}
\label{def:genmono}
A function $f(S)$ mapping a set in $2^\Yspace$ to $\mathbb{R}$ is \textit{monotone} if $\forall S, S'$, $S\subseteq S'\implies f(S)\leq f(S')$.
\end{definition}
\vspace{-2mm}
\cref{def:genmono} is very common for value functions in game theory (e.g. in fair allocation)~\cite{value1,value2}. 
The monotone requirement can be removed without affecting our general algorithmic procedures, but it keeps the cost-value trade-off more intuitive. 

For cost control, there could be two possible goals: Controlling the \textbf{expected cost}, or controlling the probability of a \textbf{cost violation}, which we will henceforth refer to as ``Expected Cost Control'' and ``Violation Control'' respectively:
\begin{definition}\label{problem:expectedcost}
(Expected Cost Control)
Given a target cost $c$, the expected cost control problem is:
\vspace{-3mm}
\begin{aligncustomsize}
    \max_{S_{N+1}\in 2^{\mathcal{Y}}} \hspace{2mm} & \mathbb{E}[V(S_{N+1};Y_{N+1})]\\
    \text{s.t.} \hspace{2mm}& \mathbb{E}[C(S_{N+1};Y_{N+1})] \leq c \label{eq:problem:expectedcostcontrol}
\end{aligncustomsize}
\vspace{-7mm}
\end{definition}
\begin{definition}\label{problem:violation}
(Violation Control)
Given a target cost $c$ and a violation probability target $\delta$, the violation control problem is defined as:
\vspace{-3mm}
\begin{aligncustomsize}
    \max_{S_{N+1}\in 2^{\mathcal{Y}}} \hspace{2mm} & \mathbb{E}[V(S_{N+1};Y_{N+1})]\\
    \text{s.t.} \hspace{2mm}& \mathbb{P}\{C(S_{N+1};Y_{N+1}) > c\} \leq \delta
\end{aligncustomsize}
\end{definition}
\vspace{-4mm}
\textbf{Remarks}:
Our notion of violation control (\cref{problem:violation}) looks similar to that of risk control defined in~\cite{ltt,bates-rcps}. 
However, apart from the additional goal to maximize value, there is a crucial difference: The equivalent of ``cost'' in their case is an \textit{expectation} on the \textit{population}, whereas \cref{problem:violation} bounds the cost on one instance (which has a large variance).
\cref{problem:expectedcost,problem:violation} can also be viewed as generalizations of~\cite{reffp} to general cost functions (instead of false positives), along with the additional value-maximization goal. 

\subsection{Examples of Cost and Value Function}\label{sec:problem:hccexample}
Next, we present concrete examples of cost and value functions to ground our discussion.
The most straightforward examples for value and cost could  be the number of True or False Positives in the prediction set.
Formally, we have:
\vspace{-2mm}
\begin{aligncustomsize}
    V_{TP}(S;y) &\defeq |\{k: k\in S, k\in y\}|\\
    C_{FP}(S;y) &\defeq |\{k: k\in S, k\not\in y\}|.
\end{aligncustomsize}
\vspace{-6mm}\\
Here, both $V_{TP}$ and $C_{FP}$ takes values in $\{0,1,\ldots,K\}$. 
In many applications, however, this might not be enough.

\def \TabNotation{
\begin{table}[ht]
\vspace{-5mm}
\caption{Notations used in this paper.}
\centering
\begin{small}
   \resizebox{1\columnwidth}{!}{
\begin{tabular}{cl}
\toprule
Symbol & Meaning \\
\midrule
$X,Y,Z$ & Random input $X$ and label set $Y$; $Z=(X,Y)$.\\
$\Yspace = [K]$ & The set of all $K$ classes\\
$\phat$, $\mathbf{v}$, $\mathbf{r}$ v.s. $\hat{p}_k$, $v_k$, $r_k$ & vector v.s. its $k$-th entry\\
\midrule
$S \in 2^\Yspace$ & Prediction set\\
$\Ucal(X)\subseteq 2^\Yspace$ & Universe to choose the prediction sets from\\
$V(S;Y)$ & Value of set $S$ (to maximize)\\
$C(S;Y)\in[0,C_{max}]$ & Cost of set $S$ (to control)\\
$\hat{C}(S;X)$, $\hat{V}(S;X)$& Cost and value proxy (estimates for $C$ and $V$)\\
$C^+(t;Z)$ & Maximum cost with proxy $\hat{C} < t$ \\
$\phat(x)=[\hat{p}_k(x)]_{k=1}^K$ & Probability prediction by the base classifier\\
\bottomrule
\end{tabular}}
\end{small}
\label{table:notation_table}
\vspace{-2mm}
\end{table}
}
\TabNotation

\textbf{Example in hospital coding}:  
For a hospital quality-assurance system detecting missing diagnoses, $S$ could be a set of disease codes.
Diseases are typically grouped into hierarchical condition categories (HCC) codes, and assigned morbidity scores (risk adjustment factors).
Some diseases are obviously more severe than others.
As illustrated in \cref{fig:hccexample}, Metastatic Cancer and Acute Leukemia is 25+ times more severe than Diabetes without Complication\footnote{According to the 2023 Midyear ``Community NonDual Aged'' factors released on \url{www.cms.gov}. }. 
The severity can be encoded in $V$ or $C$ via weights $w_k$:
\vspace{-2mm}
  \begin{small}
  \begin{align}
    V_{TPC}(S;y) &\defeq \sum_{k\in S}\mathbbm{1}\{y_k = 1\} w_k\\
    C_{FPC}(S;y) &\defeq \sum_{k\in S}\mathbbm{1}\{y_k = 0\} w_k.
  \end{align}
  \end{small}
  \vspace{-2mm} 
\vspace{-3mm}\\
We refer to $V_{TP\underline{C}}$ and $C_{FP\underline{C}}$ as ``\underline{c}ontinuous'' functions\footnote{Their ranges are technically discrete, but close to continuous for any practical value of $K$:
There are 35 codes in \dataMIMICThree, so a \texttt{float32} number cannot encode all values of $V_{TPC}$ and $C_{FPC}$.}.
$C_{FPC}$ also has applications in health insurance plans and could be interpreted as incorrectly billed conditions, as $w_k$ roughly corresponds to expected cost.
\vspace{-2mm}

\section{Methodology}\label{sec:method}
\vspace{-2mm}
In this section, we lay out the exact methodology and the attendant theory for cost control, using ideas from conformal prediction.
We will also describe how to improve the value of the final chosen prediction set.
The computational details for efficient online updates are deferred to \cref{sec:efficientupdate}.
The list of major notations can be found in \cref{table:notation_table}.
\vspace{-2mm}

\subsection{Preliminary: Split Conformal False Positive Control}\label{sec:method:prelimsplitcp}
We begin by introducing (split) conformal prediction and showing how to use it to conservatively control $C_{FP}$. 
However, we first need to define the notion of a quantile function:
\begin{definition} (Quantile Function $Q$)
    The $\beta$-quantile for a set of values $\{v_i\}_{i=1}^{N}$ is defined by the empirical CDF $F$:
\vspace{-3mm}
\begin{aligncustomsize}
    \hspace{-2mm}Q(\beta; F) \defeq \inf\{t:F(t)\geq\beta\}
    \text{ where } F(t) \defeq\sum_{i=1}^{N} 
    \frac{\Delta_{v_i}(t)}{N}\hspace{-1.5mm}\label{eq:quantilefunction}
\end{aligncustomsize}    
\end{definition}
\vspace{-5mm}
Here, $\Delta_{v}(t)$ is the indicator function $\mathbbm{1}\{t\geq v\}$.
When possible, we write $Q(\beta; \{v_i\}_{i=1}^{N})$ for readability.
\begin{theorem} (Split Conformal Prediction \cite{papadopoulos2002}) \label{thm:splitcp:general}
Assuming exchangeability of $[Z_1,\ldots,Z_{N+1}]$, for a fixed measure $g$ and any $\epsilon\in (0,1)$:
\vspace{-2mm} 
\begin{footnotesize}
\begin{align}
    \mathbb{P}\Big\{g(Z_{N+1})\leq Q\Big(1-\epsilon; \{g(Z_i)\}_{i=1}^N\cup\{\infty\}\Big)\Big\}\geq 1-\epsilon \label{eq:splitcp}
\end{align}
\end{footnotesize}
\end{theorem}
\vspace{-2mm}
\vspace{-2mm} 
\cref{eq:splitcp} can be better understood as a statement on \textit{rankings}. That is, due to exchangeability, the ranking of $g(Z_{N+1})$ among $\{g(Z_i)\}_{i=1}^{N+1}$ follows a uniform distribution.

Suppose for a moment that our goal is to control the false positive rate for class $k$ only.
We could focus on only the sub-population where $k\not\in Y$. 
If we define the \textit{nonconformity score} for $Z_i$ as $g(Z_i)\defeq p_k(X_i)$, 
\cite{Lei2014ClassificationConfidence} gives us:
\begin{theorem}\label{thm:splitcp:binary}
    Assuming $[Z_1,\ldots,Z_{N+1}]$ are i.i.d., let
    \vspace{-1mm}
    \vspace{-2mm} 
    \begin{footnotesize}
    \begin{align}
        &k\in \hat{S}_{N+1} \Leftrightarrow p_k(x_{N+1}) > t_k\\
    \text{s.t. } & t_k=Q(1-\epsilon; \{p_k(x_i): y_{i,k}=0\}\cup\{\infty\}).
    \end{align}
    \end{footnotesize}
    \vspace{-5mm}\\
    Then, we have $\mathbb{P}\{k\in \hat{S}_{N+1}|k\not\in Y_{N+1}\} \leq \epsilon$.
\end{theorem}
\vspace{-2mm}
\cref{thm:splitcp:binary} can be used to conservatively control $C_{FP}$:
\begin{theorem}(The \baselineClasswise method)\label{thm:splitcp:classwise}
    Assuming $[Z_1,\ldots,Z_{N+1}]$ are i.i.d., let $\epsilon=\frac{c}{K}$ and $S_{N+1}$ be defined like in \cref{thm:splitcp:binary}, then $\mathbb{E}[C_{FP}(\hat{S}_{N+1}, Y_{N+1})] \leq c$.
\end{theorem}
We defer all proofs to the Appendix. 
The intuition for \baselineClasswise is that we evenly distribute the ``cost budget'' to each class, so in expectation the total cost $\leq c$. 
While \cref{thm:splitcp:classwise} does the job to control $C_{FP}$, it is conservative as long as $\exists k, \mathbb{P}\{k\in Y\} > 0$. 
Moreover, it cannot handle other cost functions and lacks a notion of value.
Next, we will present \methodname to overcome these limitations.

\vspace{-1mm}
\subsection{Proxy Functions}\label{sec:method:proxy}
In \cref{sec:method:prelimsplitcp}, we chose a threshold $t_k$ for $1-p_k(X)$ based on $\{Z_{i}\}_{i=1}^N$ to make a prediction for $X$.
For more general cost control problems, we can find a proxy (or estimate) $\hat{C}$ for our cost function $C$.
In many cases, $\hat{C}$ can be derived from the base prediction $\phat$. 
For instance, for $C_{FP}$ and $C_{FPC}$, we could have the following cost proxies:
\vspace{-2mm}
\begin{aligncustomsize}
    \hat{C}_{FP}(S;x) &\defeq \sum_{k\in S} (1-\hat{p}_k(x))\label{eq:chat:fp}\\
    \hat{C}_{FPC}(S;x) &\defeq \sum_{k\in S} (1-\hat{p}_k(x)) w_k .\label{eq:chat:fpc}
\end{aligncustomsize}
\vspace{-3mm}\\
For a generic, non-additive set function $C$, one possible proxy could be a Monte-Carlo estimate:
\vspace{-1.5mm}
\begin{aligncustomsize}
    \hat{C}_{MC}(S;x) &\defeq \hat{\mathbb{E}}_{y\sim \phat(x)}[C(S; (x,y))]\label{eq:chat:montecarlo}.
\end{aligncustomsize}
\vspace{-5mm}\\
The value proxy $\hat{V}$ can be defined analogously to $\hat{C}$.
Note that the sampling in \cref{eq:chat:montecarlo} is trickier than it might seem because we cannot derive class correlations from $\phat$. 
One could also potentially estimate such correlational information, or directly train a $\hat{C}$ on the training set.
Next, the cost proxy will be used to select the acceptable prediction sets.

\subsection{The Expected Cost Control}\label{sec:method:expectedcost}
To proceed, we fix a $\Ucal(X)\subseteq 2^\Yspace$, the universe of all label sets that we would want to consider.
For example, when $\Yspace$ is small, we could simply enumerate $\Ucal(X)=2^\Yspace$.
The choice of $\Ucal$ is discussed in \cref{sec:method:universe}.
For both expected cost and violation control, we will create a total ordering of the subsets of $\Ucal$, which allows us to pick a threshold like before to keep the ``acceptable'' sets. 
The rest of section \ref{sec:method}, from here, elucidates these details and theoretical guarantees.

Define the maximum cost corresponding to a threshold $t$ as:
\vspace{-5mm}
\begin{aligncustomsize}
C^+(t;z) &\defeq \max_{S\in \Ucal(x)}\{C(S;z): \hat{C}(S; x) < t\}.\label{eq:maxcost}
\end{aligncustomsize}
\vspace{-3mm}\\
Now, we create a total ordering via $C^+$, allowing us to pick a threshold $T_c$ to control $C^+$ (thus $C$):
\vspace{-2mm}
\begin{aligncustomsize}
    T_c &= \sup\Big\{t: \frac{C_{max} + \sum_{i=1}^N C^+(t;Z_i)}{N+1}\leq c\Big\}\label{eq:expected:threshold}
\end{aligncustomsize}
\vspace{-3mm}\\
Finally, we simply pick the expected value-maximizing set:
\vspace{-5mm}
\begin{aligncustomsize}
    \hat{S}_{N+1} &\defeq \argmax_{S\in \Ucal(x_{N+1}): \hat{C}(S; x_{N+1}) < T_c} \hat{V}(S;x_{N+1}). \label{eq:chooseoutput:expectation}
\end{aligncustomsize}
\vspace{-4mm}\\
We get a cost control guarantee similar to that in~\cite{reffp}, first proven in~\cite{riskcontrolproof}:
\begin{theorem}\label{thm:control:expectation}
Assuming exchangeable $\{Z_i\}_{i=1}^{N+1}$.
For any $c\in(0,C_{max}]$, and $\hat{S}_{N+1}$ from \cref{eq:chooseoutput:expectation}:
\vspace{-2mm}
\begin{aligncustomsize}
    \mathbb{E}_{Z_1,\ldots,Z_{N+1}}[C(\hat{S}_{N+1}; Z_{N+1})] \leq c
\end{aligncustomsize}
\end{theorem}
\vspace{-5mm}

\subsection{The Violation Control}\label{sec:method:violation}
Analogously to the above, for the violation control, we could have the threshold $T_{c,\delta}$ and prediction set $\hat{S}_{N+1}$ as: 
\vspace{-2mm}
\begin{aligncustomsize}
    T_{c, \delta} &= \sup\Big\{t: \frac{\sum_{i=1}^N \mathbbm{1}\{C^+(t;Z_i)\leq c\}}{N+1}\geq 1- \delta \Big\}\label{eq:violation:threshold},\\
    \hat{S}_{N+1} &\defeq \argmax_{S\in \Ucal(X_{N+1}): \hat{C}(S; X_{N+1}) < T_{c,\delta}} \hat{V}(S;X_{N+1}).\label{eq:chooseoutput:violation}
\end{aligncustomsize}
\vspace{-8mm}\\
\begin{theorem}\label{thm:control:violation}
Assuming exchangeable $\{Z_i\}_{i=1}^{N+1}$.
For any $c\in (0, C_{max}]$, $\delta\in(0,1)$, and $\hat{S}_{N+1}$ from \cref{eq:chooseoutput:violation}: 
\vspace{-2mm}
\begin{aligncustomsize}
    \mathbb{P}_{Z_1,\ldots,Z_{N+1}}\{C(\hat{S}_{N+1}; Z_{N+1}) \leq c\} \geq 1-\delta
\end{aligncustomsize}
\end{theorem}
\vspace{-5mm}
Note that the randomness is taken over $Z_1,\ldots,Z_{N+1}$.
In particular, it means the guarantees hold with respect to a random $X_{N+1}$, and is only marginal. 
The proofs for both \cref{thm:control:expectation} and \cref{thm:control:violation} (see Appendix) simply leverage the exchangeability of $\{Z_i\}_{i=1}^{N+1}$. 
\vspace{-4mm}

\subsection{Value-Maximization via the Choice of $\Ucal$}\label{sec:method:universe}
The choice of $\Ucal$ significantly impacts performance. 
To see why, suppose $\Ucal(X)$ is an uninformative random subset of $2^\Yspace$ that does not depend on $X$.
Since we only pick one element $\hat{S}\in\Ucal$, but perform the cost control on the entire $\{S\in \Ucal: \hat{C}(S; X) \leq T\}$ (see \cref{eq:chooseoutput:expectation}), the gap between $C^+$ and $C(\hat{S})$ will be larger as $|\Ucal|$ increases, leading to more conservative $\hat{S}$.
On the flip side, as $\Ucal$ expands, it also could potentially include a set with higher expected value.
Using the information from $X$ (and our knowledge of the value and cost functions) could potentially achieve higher value while satisfying cost control requirements. 

\textbf{The Full Universe}
A simple starting point, as suggested earlier, is $\Ucal(X) = 2^\Yspace$.
We denote this as $\Ucal_{full}$.
Clearly, $\Ucal_{full}$ is only practical when $K$ is small, as $|\Ucal| = 2^K$.
We consider more practical universes $\Ucal$ below. 

As many real-world value/cost functions of interest are additive, we discuss the additive case before extending to the general case.

\textbf{The Greedy Sequence (Additive)}
When $K$ is large, we could employ a different strategy and add one class at a time and create a sequence of sets, by having $\Ucal(x) = \{S_{(j)}(x)\}_{j=0}^{K}$ with $S_{(0)}(x)=\emptyset$ and 
\vspace{-2mm}
\begin{aligncustomsize}
    S_{(j+1)}(x) &= S_{(j)}(x)\cup \{\sigma(j+1;x)\}\label{eq:univ:seq:inductive}
\end{aligncustomsize}
\vspace{-4mm}\\
where $\sigma$ is an ordering of $[K]$ that depends on $x$.
A recent work on \textit{false positive control}~\cite{reffp} proposes to greedily add classes with the highest predicted probability:
\vspace{-2mm}
\begin{aligncustomsize}
    \sigma_{\baselineGreedyProbSuffix}(x) &\defeq argsort(-\phat(x))\label{eq:order:greeedyprob}
\end{aligncustomsize}
\vspace{-5mm}\\
This happens to be suitable given their particular goal: Maximizing the true positive rate (TPR).
To extend this idea, it might seem natural to simply replace the TPR-maximizing sequence with the value-maximizing sequence:
\vspace{-2mm}
\begin{aligncustomsize}
    \sigma_{\baselineGreedyValSuffix}(x) &\defeq argsort(-\phat(x)\odot\mathbf{v})\label{eq:order:greedyvalue}
\end{aligncustomsize}
\vspace{-5mm}\\
where $\mathbf{v}$ denotes the value vector, with $v_k$ being the value for class $k$, and $\odot$ being the Hadamard product. 
However, this is sub-optimal, because it fails to consider the trade-off between value and cost. 
Instead, we propose the \textbf{ratio-maximizing sequence}, where the ratio refers to the marginal value divided by the marginal cost proxy:
\vspace{-2mm}
\begin{aligncustomsize}
    \sigma_{\baselineGreedyRatioSuffix}(x) &\defeq argsort(-\mathbf{r})
    \text{ s.t. } r_k\defeq \hat{p}_k(x) v_k/\hat{C}_k(x).\label{eq:ratio:additive}
\end{aligncustomsize}
\vspace{-5mm}\\
Here, $\hat{C}_k(X)$ denotes the predicted marginal cost for class $k$.
For example, for $\hat{C}_{FP}$ (\cref{eq:chat:fp}), $\hat{C}_{FP,k}(x)=1-\hat{p}_k(x)$.
Note that we dropped the dependency on $S$ here because of the additive assumption. 

\textbf{The Greedy Ratio Sequence (General)}
The \baselineGreedyRatioSuffix method described above could be further generalized to any cost and value functions:
\vspace{-2mm}
\begin{aligncustomsize}
    & \sigma'_{\baselineGreedyRatioSuffix}(j+1, x) \defeq \argmax_{k\not\in S_{(j)}(x)} r(k, S_{(j)})\\
    \text{where } & r(k, S)\defeq \frac{\hat{V}(S\cup\{k\}) - \hat{V}(S)}{\hat{C}(S\cup \{k\}) - \hat{C}(S)}.\label{eq:ratio:general}
\end{aligncustomsize}
\vspace{-4mm}\\
\textbf{
\methodname by default uses \cref{eq:ratio:general}}, as well as the faster special case \cref{eq:ratio:additive} when possible.
\vspace{-3mm}
\section{Fast Online Update with \datastructname}\label{sec:efficientupdate}
We have presented the analytical form for the thresholds for both the expected cost and violation control in \cref{eq:expected:threshold} and \cref{eq:violation:threshold}.
However, both involve computing the supremum over a set of feasible values, which is costly, especially in an online setting where the thresholds need to be constantly re-computed. 
In this section, we will first cast the threshold-finding problem into a more general task of quantile finding, and then present an efficient implementation for the quantile finding problem (thus permitting computation of the thresholds dynamically).
These two tricks reduce the complexity by a factor of $N$ (sample size). 
We focus on computing $T_c$ and defer $T_{c,\delta}$ to the Appendix as the latter is much simpler. 

Suppose each sample corresponds to up to $M$ ($=|\Ucal(X)|$) thresholds.
There are $O(MN)$ potential thresholds.
For each threshold $t$, evaluating $\sum_{i=1}^N C^+(t;Z_i)$ is $\Omega(MN)$ depending on the implementation.
If we perform a binary search on the thresholds, \textit{each online update} will cost at least $O(MN\log{(MN)})$, or roughly $O(N\log{N})$ when $N\gg M$.
This is too expensive when $N$ is large - for example, tens of billions for a company processing insurance claims. 

The first step towards an efficient algorithm is the observation that finding the threshold is similar to finding a weighted quantile of all thresholds recorded so far.
Formally, given a set of values $v_1,\ldots,v_{N+1}$ with weights $w_1,\ldots,w_{N+1}$, the CDF $F$ of the weighted distribution is~\cite{localguan22}:
\vspace{-2mm}
\vspace{-1mm} 
\begin{aligncustomsize}
    F(t)\defeq\sum_{i=1}^{N+1} w_i \Delta_{v_i}(t) \text{ where } \sum_{i=1}^{N+1} w_i = 1.
\end{aligncustomsize}
\vspace{-1mm} 
\vspace{-3mm}\\
Recall the definition of $Q$ as was provided in \cref{eq:quantilefunction}. Now, we formally establish the equivalence between $T_c$ and a weighted quantile, in the theorem below:
\begin{theorem}\label{thm:equivalence:quantile}
For each data $x_i$, sort $\Ucal(x_i)$ as $[S_{i,(1)},\ldots,S_{i,(M)}]$ such that $\hat{c}(S_{i,(j)})\leq \hat{c}(S_{i,(j+1)})$.
Assuming $\forall x$, $\emptyset\in \Ucal(x)$ and $C(\emptyset;Z)=\hat{C}(\emptyset;X) = 0$.
Let
\vspace{-4mm} 
\vspace{-1mm}
\begin{aligncustomsize}
    c^+_{i,(j)} &= \max_{j'\leq j}\{C(S_{i,(j')};z_i)\}\\
    w_{i,(j)} &= \Big(\frac{c^+_{i,(j)} - c^+_{i,(j-1)}}{W} \Big)\mathbbm{1}\{j>1\}.
\end{aligncustomsize}
\vspace{-4mm}\\
where $W=\sum_{i,j} w_{i,j}$ is a normalizing factor.
Denote the weighted quantile as 
\vspace{-1mm}
\vspace{-2mm} 
 \begin{aligncustomsize}
     T'_c \defeq Q(\frac{c(N+1)-C_{max}}{W}; F=\sum_{i=1}^N\sum_{j}w_{i,(j)}\Delta_{\hat{c}_{i,(j)}})
 \end{aligncustomsize}
 \vspace{-2mm} 
\vspace{-3mm}\\
 Then, $T_c$ is the immediate successor of $T'_c$ when $\frac{c(N+1)-C_{max}}{W}$ is in the domain of $F$ in $\{\hat{c}(S_{i,(j)})\}_{i,j}$ (unlikely in practice).
 Otherwise, $T'_c = T_c$.
\end{theorem}
\vspace{-2mm} 
With \cref{thm:equivalence:quantile}, the problem becomes finding the weighted quantile\footnote{Strictly speaking, we sometimes need to find its immediate successor. However, the chance of this is very low, and if we skip this step, we are only changing $\leq$ to $<$ in \cref{eq:problem:expectedcostcontrol}.} in an efficient manner. 
We will achieve this with a custom data structure, and our high-level idea is to combine a self-balancing binary search tree and a segmentation tree: The former allows for efficient updates of $\{c_{i,(j)}\}_{j=1}^M$, and the latter enables efficient query of the weighted quantile.
The \texttt{INSERT} and \texttt{QUERY} operations are briefly described in \cref{alg:main:tree}.
We refer to this data structure as \datastructname. 
To the best our knowledge, this is a new data structure to store and query any empirical CDF and is perhaps of independent interest\footnote{Much effort has been denoted to estimating quantiles in very large-scale applications, but here we focus on the \textit{exact} quantiles of any discrete CDF. See discussion in \cref{sec:relatedworks}.}. 
Several technical details are deferred to the Appendix, including the (simpler) computation of $T_{c,\delta}$ and how to \texttt{DELETE} data points, which is important if in practice we need to account for distribution shift with a rolling (instead of expanding) window in calibrating our thresholds. 
Complete implementations are included in the supplementary material. 
Finally, we present the full pipeline of \methodname for expected cost control in \cref{alg:main:expected}, with the slightly different version for violation control in the Appendix.

\def \AlgMainWeightedQuantile{
\vspace{-1mm} 
\vspace{-2mm}
\begin{algorithm}[h]
\SetAlgoLined
\DontPrintSemicolon
\SetKwProg{Struct}{Struct}{:}{}
\Struct{Node}{
\texttt{color} \# For red/black marking\\
\texttt{value} \# Recording $v_i$ (the threshold in $\Delta_{v_i}$)\\
\texttt{sum} \# The sum of all $w_i$ under this node\\
\texttt{weight} \# $w_i$. \texttt{weight} (thus \texttt{sum}) is un-normalized.
}
\SetKwFunction{Finsert}{INSERT}
\SetKwProg{Fn}{Function}{:}{}
\Fn{\Finsert{value $v$, weight $w$}}{
    Insert a leaf Node(\texttt{value}=$c$,\texttt{weight}=$w$)\\
   Fix the tree to a valid red-black tree, while maintaining valid \texttt{sum} values.
}
\textbf{End Function}

\SetKwFunction{Fquery}{QUERY}
\Fn{\Fquery{quantile $q$}}{
    Compute the cumulative weight $cw=$\texttt{root.sum}$*q$.\\
    Find the \texttt{node}($cw$) corresponding to $cw$ using binary search on \texttt{sum}.\\
    \textbf{return} \texttt{node}$(cw)$\texttt{.value}
}
\textbf{End Function}
\caption{\datastructname (short version)}
\label{alg:main:tree}
\end{algorithm}
\vspace{-2mm}
\vspace{-3mm} 
}
\AlgMainWeightedQuantile

\def \AlgMainExpectedCost{
\begin{algorithm}[ht]
\SetAlgoLined
\DontPrintSemicolon
\begin{algorithmic}
   \STATE {\bfseries Input:} data $\{(X_1,Y_1), (X_2,Y_2), \ldots\}$, value and cost function $V$ and $C$
   \STATE \texttt{tree} $\gets$ a \datastructname object.
   \FOR{$N=0, 1, \ldots$}
   \IF{$N > $ some burn-in period}
   \STATE $T_c\gets$ \texttt{tree.QUERY}($\frac{(N+1)\cdot c-C_{max}}{tree.root.sum}$)
   \STATE Find the \textbf{greedy-ratio} $\Ucal(x_{N+1})$ with \cref{eq:univ:seq:inductive,eq:ratio:general} and the relevant proxy $\hat{V}$ and $\hat{C}$ from \cref{sec:method:proxy}.
   \STATE Pick $\hat{S}_{N+1}$ from $\Ucal(x_{N+1})$ with \cref{eq:chooseoutput:expectation} and $T_c$.
   \ENDIF
   \STATE Insert $\{(\hat{c}_{N,(j)}, w_{N,(j)})\}_{j=1}^M$ to \texttt{tree} (\cref{thm:equivalence:quantile}).
   \ENDFOR
\end{algorithmic}
\caption{Expected Cost Control (online)}
\label{alg:main:expected}
\end{algorithm}
\vspace{-3mm} 
}
\AlgMainExpectedCost

\textbf{Complexity}:
The sorting of $[c_{i,(j)}]$ takes $O(M\log{M})$ and each insertion into the tree takes $O(\log{(MN)})$. 
Thus, inserting a data point $(x_i, y_i)$ takes $O(M\log{(MN)})$ in total (same for deletion).
A query step takes $O(\log{(MN)})$.
They add up to $O(M\log{(MN)})$ per update, which improves at least a factor of $N$ over the binary search version with a complexity of $O(MN\log{(MN)})$ (or $O(\log{N})$ vs $O(N\log{N})$ if we consider $N\gg M$). 

\commentZL{
Note that in the analysis above, we did not try to link $M$ to the number of classes $K$, as this depends heavily on the choice of $\Ucal$. 
Sometimes enumerating $\Ucal$ could be very expensive, up to $2^K$. 
However, in real-world scenarios usually $N$ grows to $\gg M$, so eventually the order of $M$ is less important.
}

\vspace{-3mm} 

\section{Experimental Evaluation}\label{sec:exp}

\subsection{Datasets and Tasks}
We consider the following datasets and tasks for evaluation.
\vspace{-7mm}
\begin{itemize}
    \item \dataMNIST: 
    A synthetic dataset created by superimposing MNIST~\cite{mnist} images, with each class appearing with a probability of $0.4$. 
    The task is to detect all digits that appear in the image. 
    Training/test split is 54000/10000.\vspace{-2mm}   
    
    \item \dataMIMICThree:  
    The MIMIC-III dataset~\cite{MIMIC,PhysioNet,MIMICDemo} is collected from the critical care units of the Beth Israel Deaconess Medical Center and contains data from 38K patients. 
    For each discharge report, we randomly mask out 30\% of the HCC codes (and the associated ICD-9 codes) and remove the relevant diagnoses section in the discharge notes.
    Input include the remaining free-text notes and patient features pre-processed by~\cite{mimic3preprocessing}.
    Training/test split is 27,895/8,570.\vspace{-2mm}

    \item \dataClaim: 
    Insurance claim prediction on a proprietary dataset from a large healthcare data provider in North America. 
    For each claim date, we use the previous 26 weeks to predict any HCC code to appear in the next 26 weeks.
    Training/test split is 236,089/75,484 .\vspace{-2mm}   
    
\end{itemize}
For both \dataMIMICThree and \dataClaim, we created a ``(select)'' version that only focuses on 10 frequent codes in order to evaluate $\Ucal_{full}$ (which requires enumerating all $2^K$ sets).
We create the synthetic \dataMNIST to evaluate different methods on more complex (made-up) value functions, as well as when the label is not too sparse. ($\mathbb{E}[|Y|]$ ranges from 0.4 to 0.8 in \dataMIMICThree and \dataClaim). 


\textbf{Value and Cost Functions}
For \dataMIMICThree and \dataClaim, we define $V_{TP}$ and $C_{FP}$ like in \cref{sec:problem:hccexample}, as well as $ V_{TPC}$ and $C_{FPC}$ with the corresponding risk adjustment factors.
We let $w_k = k$ and define the same for \dataMNIST\footnote{We treat class $0$ as $10$ so it has nonzero weight.}.
To illustrate the flexibility of \methodname, we also define a ``general'' (non-additive) value function  for MNIST:
\vspace{-2mm}
\begin{aligncustomsize}
    V_{GEN}(S;Y) &= \prod_{k\in S\cap Y} \frac{k+5}{10} + \sum_{k\in S\cap Y} (k-5)^2.
\end{aligncustomsize} 
\vspace{-2mm}\\
For readability, all value and cost functions have been normalized to take values in $[0,100]$. 

\textbf{Experiment Setup}
We set the target cost $c$ from 5 to 50 with a stride of 5 (rounded for $C_{FP}$), as well as the commonly used $\delta=0.1$. 
The base classifiers are deep neural networks, with architecture and training details in the Appendix.
For $C_{FP}$ control, we use the continuous values functions.
 For $C_{FPC}$ control, we use all value functions: $V_{TP}, V_{TPC}$ and $V_{MULT}$.
$V_{TP}$ with $C_{FP}$ is skipped because it reduces to experiments in~\cite{reffp}.
For each random seed, we randomly sample 3000 samples from the test set and use 1000 points as ``burn-in''. 
Mean and standard deviations for 10 seeds are reported.
For \dataClaim, we include a ``sequential'' version (\dataClaimSeq) in the Appendix with a temporal train/test split of 75/25 to simulate real-world applications.

\subsection{Baselines}
We compare our method with the following baselines (where applicable, as none can be applied to all experiments):
\baselineInnerSet, a one-sided variant of \cite{InnerSet} adapted to FP control in~\cite{reffp}; 
\baselineClasswise, which controls class-wise false positive with $\epsilon=\frac{c}{K}$ (\cref{thm:splitcp:classwise});
\baselineFPCP~\cite{reffp}, equivalent to \cref{eq:order:greeedyprob} but trains an additional DeepSets~\cite{DeepSets} as $\hat{C}_{FP}$.
All these methods do not support flexible value functions, but where applicable they should provide (sometimes conservative) cost control.

We also include variants of \methodname with different $\Ucal$: \baselineGreedyValSuffix (\cref{eq:order:greedyvalue}),  \baselineGreedyProbSuffix (\cref{eq:order:greeedyprob})  and \baselineFullSuffix (using $\Ucal_{full}$).
Unless specified, \methodname refers to the variants with greedy ratio (\cref{eq:ratio:general,eq:ratio:additive}).
We want emphasize again that in controlling $C_{FP}$, \baselineGreedyProb behaves exactly the same as \baselineFPCP.
However, we refer to it as \baselineGreedyProbSuffix in $C_{FPC}$ control, as \baselineFPCP does not apply to general cost functions and technically is too costly to compute in its original form (see \cref{fig:main:exp:time}).
This name also highlights the cause of different performance  -  difference in $\Ucal$.

\textbf{Evaluation Metrics} include the empirical mean of Excess Cost ($C(\hat{S})-c$), Violation Frequency ($\mathbbm{1}\{C(\hat{S}) > c\}$), and Value ($V(\hat{S})$), as well as Computation Time.

\def \FigureTime{
\begin{figure}[ht]
\vskip -0.1in
\centering
\centerline{\includegraphics[width=0.8\columnwidth]{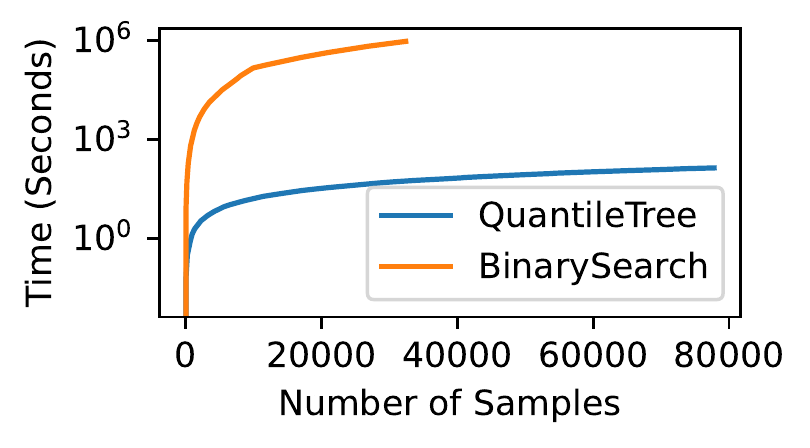}}
\vspace{-5mm}
\caption{
Computation time as a function of $N$, on \dataClaimSeq.
As expected, \datastructname gracefully handles large $N$, while BinarySearch does not finish.
}
\label{fig:main:exp:time}
\vskip -0.2in
\end{figure}
}
\FigureTime

\def \FigureCurve{
\begin{figure}[ht]
\centering
\centerline{\includegraphics[width=\columnwidth]{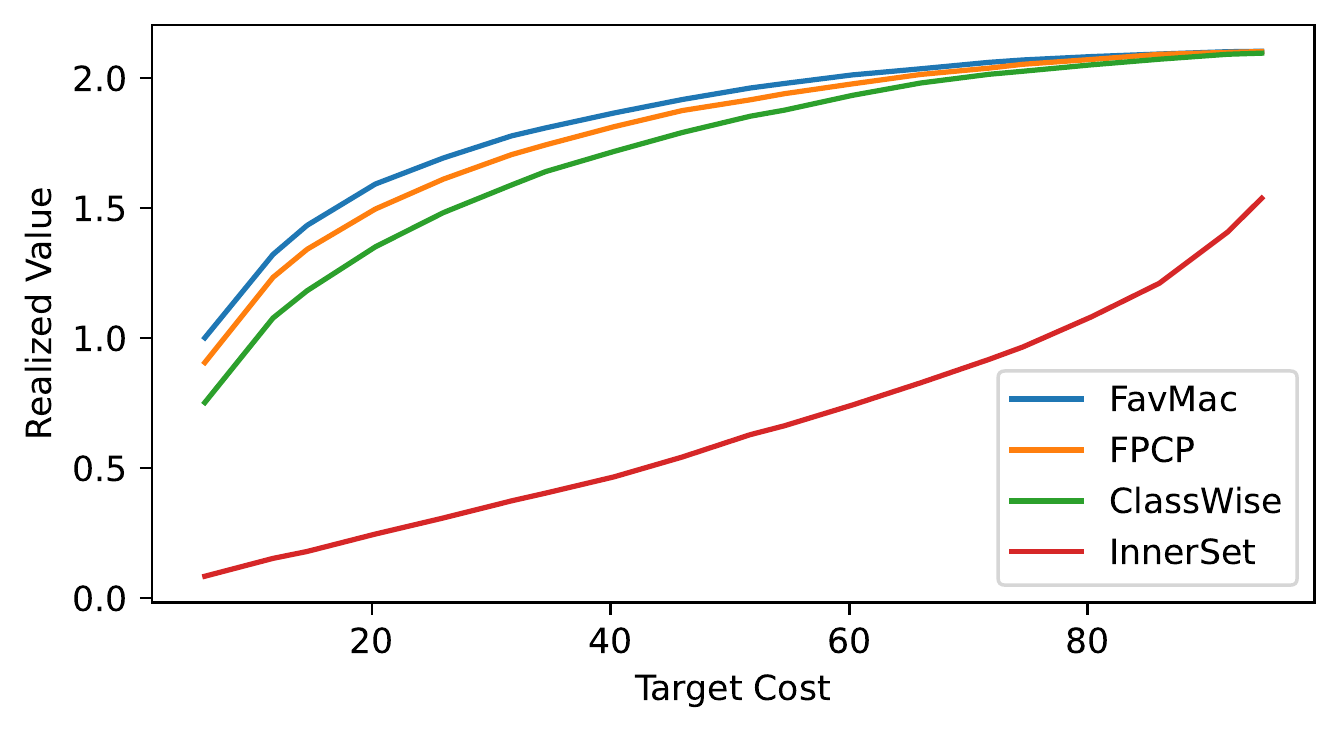}}
\vskip -0.2in
\caption{
The $V_{TPC}$ vs $C_{FP}$ (value-cost) curve on \dataMIMICThree. 
Here we extend the target cost up to 95.
\methodname achieves better trade-offs than \baselineFPCP, while \baselineInnerSet and \baselineClasswise are conservative, realizing lower values. 
}
\label{fig:main:exp:tradeoffcurve}
\vskip -0.1in
\end{figure}
}
\FigureCurve

\def \TableMainFPValueExpectation{
\begin{table}[ht]
\vspace{-3mm}
\caption{
Mean value of the prediction sets with \textit{expected} false positive control.
The higher the better. 
}
\label{table:main:exp:value:FP:expectedcost}
\vskip 0.05in
\centering
\begin{small}
   \resizebox{1\columnwidth}{!}{
\begin{tabular}{lccc|c}
\toprule
Value ($\uparrow$)
 & \baselineClasswise & \baselineInnerSet & \baselineFPCP & \methodname\\
\midrule
\dataMIMICThree(select) & 2.14$\pm$0.08 & 0.87$\pm$0.06 & 2.16$\pm$0.07 & \textbf{2.26$\pm$0.08}\\
\dataMIMICThree & 1.44$\pm$0.05 & 0.34$\pm$0.03 & 1.56$\pm$0.04 & \textbf{1.64$\pm$0.05}\\
\dataClaim(select) & 1.83$\pm$0.11 & 1.45$\pm$0.06 & 2.02$\pm$0.10 & \textbf{2.11$\pm$0.11}\\
\dataClaim & 1.17$\pm$0.07 & 0.68$\pm$0.04 & 1.18$\pm$0.07 & \textbf{1.22$\pm$0.06}\\
\dataMNIST & 35.34$\pm$0.23 & 24.06$\pm$0.34 & 37.44$\pm$0.27 & \textbf{38.20$\pm$0.26}\\
\dataMNIST(GEN) & 35.87$\pm$0.28 & 24.76$\pm$0.35 & 38.31$\pm$0.32 & \textbf{39.64$\pm$0.34}\\
\bottomrule
\end{tabular}
}
\end{small}
\vskip -0.25in
\end{table}
}
\TableMainFPValueExpectation

\def \TableMainFPValueViolation{
\begin{table}[ht]
\caption{
Mean value of the prediction sets with false positive \textit{violation} control.
The higher the better. 
}
\label{table:main:exp:value:FP:violation}
\centering
\vskip 0.05in
\begin{small}
   \resizebox{0.9\columnwidth}{!}{
\begin{tabular}{lcc|c}
\toprule
Value ($\uparrow$)
 & \baselineInnerSet & \baselineFPCP & \methodname\\
\midrule
\dataMIMICThree(select) & 0.30$\pm$0.04 & 2.08$\pm$0.07 & \textbf{2.21$\pm$0.07}\\
\dataMIMICThree & 0.14$\pm$0.02 & 1.41$\pm$0.04 & \textbf{1.61$\pm$0.05}\\
\dataClaim(select) & 1.02$\pm$0.05 & 2.00$\pm$0.10 & \textbf{2.09$\pm$0.11}\\
\dataClaim & 0.42$\pm$0.02 & 0.91$\pm$0.05 & \textbf{1.21$\pm$0.06}\\
\dataMNIST & 16.12$\pm$0.54 & 36.39$\pm$0.26 & \textbf{37.31$\pm$0.25}\\
\dataMNIST(GEN) & 16.84$\pm$0.57 & 37.26$\pm$0.30 & \textbf{38.80$\pm$0.34}\\
\bottomrule
\end{tabular}
}
\end{small}
\vskip -0.2in
\end{table}
}
\TableMainFPValueViolation

\def \TableMainFPCValue{
\begin{table}[ht]
\caption{
Mean value of the prediction sets with continuous cost control by variants of \methodname.
The difference is only the $\Ucal$ being used. 
The \baselineGreedyRatioSuffix version of \methodname consistently outperforms other variants.
(\baselineGreedyProbSuffix is equivalent to \baselineGreedyRatioSuffix with $V_{TPC}$).
}
\label{table:main:exp:value:FPC}
\centering
\vskip 0.05in
\begin{small}
   \resizebox{1\columnwidth}{!}{
\begin{tabular}{lccc|c}
\toprule
Expectation Control
& \baselineFullSuffix & \baselineGreedyProbSuffix & \baselineGreedyValSuffix & \baselineGreedyRatioSuffix\\
\midrule
\dataMIMICThree(select,TP) & 3.49$\pm$0.10 & \textbf{3.63$\pm$0.10} & \textbf{3.63$\pm$0.10} & \textbf{3.63$\pm$0.11}\\
\dataMIMICThree(TP) & \textendash & 1.91$\pm$0.02 & 1.91$\pm$0.02 & \textbf{1.95$\pm$0.03}\\
\dataClaim(select,TP) & 4.51$\pm$0.16 & 4.51$\pm$0.16 & 4.51$\pm$0.16 & \textbf{4.55$\pm$0.16}\\
\dataClaim(TP) & \textendash & 1.85$\pm$0.08 & 1.85$\pm$0.08 & \textbf{1.88$\pm$0.09}\\
\dataMNIST(TP) & 34.62$\pm$0.26 & 36.80$\pm$0.29 & 36.80$\pm$0.29 & \textbf{37.05$\pm$0.29}\\
\midrule
\dataMIMICThree(select,TPC) & 1.93$\pm$0.05 & 2.09$\pm$0.05 & \textbf{2.12$\pm$0.06} & 2.09$\pm$0.05\\
\dataMIMICThree(TPC) & \textendash & \textbf{1.59$\pm$0.04} & 1.50$\pm$0.04 & \textbf{1.59$\pm$0.04}\\
\dataClaim(select,TPC) & 2.05$\pm$0.08 & \textbf{2.08$\pm$0.09} & 2.02$\pm$0.10 & \textbf{2.08$\pm$0.09}\\
\dataClaim(TPC) & \textendash & \textbf{1.24$\pm$0.06} & 1.18$\pm$0.06 & \textbf{1.24$\pm$0.06}\\
\dataMNIST(TPC) & 33.87$\pm$0.22 & \textbf{36.65$\pm$0.25} & 35.99$\pm$0.26 & \textbf{36.65$\pm$0.25}\\
\dataMNIST(GEN) & 36.54$\pm$0.27 & 37.60$\pm$0.31 & 38.39$\pm$0.34 & \textbf{38.87$\pm$0.32}\\
\midrule
Violation Control\\
\midrule
\dataMIMICThree(select,TP) & 3.40$\pm$0.10 & 3.32$\pm$0.09 & 3.32$\pm$0.09 & \textbf{3.43$\pm$0.10}\\
\dataMIMICThree(TP) & \textendash & 1.84$\pm$0.03 & 1.84$\pm$0.03 & \textbf{1.89$\pm$0.03}\\
\dataClaim(select,TP) & 4.42$\pm$0.16 & 4.44$\pm$0.16 & 4.44$\pm$0.16 & \textbf{4.45$\pm$0.16}\\
\dataClaim(TP) & \textendash & 1.81$\pm$0.08 & 1.81$\pm$0.08 & \textbf{1.84$\pm$0.08}\\
\dataMNIST(TP) & 31.10$\pm$0.22 & 32.76$\pm$0.26 & 32.76$\pm$0.26 & \textbf{33.48$\pm$0.25}\\
\midrule
\dataMIMICThree(select,TPC) & \textbf{1.86$\pm$0.05} & \textbf{1.86$\pm$0.05} & 1.72$\pm$0.05 & \textbf{1.86$\pm$0.05}\\
\dataMIMICThree(TPC) & \textendash & \textbf{1.51$\pm$0.04} & 1.38$\pm$0.04 & \textbf{1.51$\pm$0.04}\\
\dataClaim(select,TPC) & 1.99$\pm$0.08 & \textbf{2.02$\pm$0.08} & 1.88$\pm$0.08 & \textbf{2.02$\pm$0.08}\\
\dataClaim(TPC) & \textendash & \textbf{1.21$\pm$0.06} & 1.14$\pm$0.06 & \textbf{1.21$\pm$0.06}\\
\dataMNIST(TPC) & 30.02$\pm$0.22 & \textbf{32.37$\pm$0.26} & 30.76$\pm$0.29 & \textbf{32.37$\pm$0.26}\\
\dataMNIST(GEN) & 33.17$\pm$0.27 & 33.46$\pm$0.30 & 34.40$\pm$0.36 & \textbf{35.65$\pm$0.33}\\
\bottomrule
\end{tabular}
}
\end{small}
\vskip -0.1in
\end{table}
}
\TableMainFPCValue

\def \TableMainFPExcessCost{
\begin{table}[ht]
\caption{
Excess cost, where values that are significant different from 0 (too conservative) are marked \costfail{red}.
}
\label{table:main:exp:excesscost:FP}
\centering
\vskip 0.05in
\begin{small}
   \resizebox{1\columnwidth}{!}{
\begin{tabular}{lccc|c}
\toprule
Excess Cost 
& \baselineClasswise & \baselineInnerSet & \baselineFPCP & \methodname\\
\midrule
\dataMIMICThree(select) & \costfail{-2.32$\pm$0.46} & \costfail{-26.58$\pm$0.13} & 0.02$\pm$0.15 & -0.01$\pm$0.15\\
\dataMIMICThree & \costfail{-0.91$\pm$0.36} & \costfail{-27.10$\pm$0.05} & -0.04$\pm$0.08 & -0.05$\pm$0.05\\
\dataClaim(select) & \costfail{-2.48$\pm$0.51} & \costfail{-23.40$\pm$0.26} & -0.03$\pm$0.19 & -0.04$\pm$0.20\\
\dataClaim & \costfail{-1.15$\pm$0.57} & \costfail{-24.40$\pm$0.19} & -0.03$\pm$0.06 & -0.05$\pm$0.05\\
\dataMNIST & \costfail{-13.14$\pm$0.25} & \costfail{-27.09$\pm$0.11} & -0.08$\pm$0.10 & -0.07$\pm$0.12\\
\dataMNIST(GEN) & \costfail{-13.14$\pm$0.25} & \costfail{-27.09$\pm$0.11} & -0.08$\pm$0.10 & -0.08$\pm$0.09\\
\bottomrule
\end{tabular}
}
\end{small}
\vskip -0.2in
\end{table}
}
\TableMainFPExcessCost

\def \TableMainFPViolation{
\begin{table}[!ht]
\caption{
Violation probability, where values that are significantly different from $\delta$ (too conservative) are marked \costfail{red}.
}
\label{table:main:exp:violation:FP}
\centering
\vskip 0.05in
\begin{small}
   \resizebox{1\columnwidth}{!}{
\begin{tabular}{lcc|c}
\toprule
Violation ($\delta=10\%$)
 & \baselineInnerSet & \baselineFPCP & \methodname\\
\midrule
\dataMIMICThree(select) & \costfail{0.13$\pm$0.02} & 10.05$\pm$0.49 & 10.13$\pm$0.32\\
\dataMIMICThree & \costfail{0.01$\pm$0.00} & 10.00$\pm$0.48 & 9.95$\pm$0.55\\
\dataClaim(select) & \costfail{0.30$\pm$0.07} & 9.94$\pm$0.69 & 9.75$\pm$0.73\\
\dataClaim & \costfail{0.06$\pm$0.02} & 10.08$\pm$0.34 & 9.84$\pm$0.45\\
\dataMNIST & \costfail{0.10$\pm$0.03} & 9.88$\pm$0.42 & 9.91$\pm$0.39\\
\dataMNIST(GEN) & \costfail{0.10$\pm$0.03} & 9.88$\pm$0.42 & 10.08$\pm$0.48\\
\bottomrule
\end{tabular}
}
\end{small}
\vskip -0.1in
\end{table}
}
\TableMainFPViolation

\subsection{Results}
Below we compare main results from the experiments, with full results in the Appendix due to space constraints.
We use p-value=0.01 for all tables (paired t-test when possible). 

The mean \textbf{Values} for false positive control are reported in \cref{table:main:exp:value:FP:expectedcost,table:main:exp:value:FP:violation}, where \methodname significantly outperforms baselines, giving up to 33\% relative improvement against the best baseline, with p-value from 7e-14 to 1e-5.
\baselineClasswise and \baselineInnerSet are fundamentally conservative, as reflected in the low value their prediction sets achieve.
\cref{table:main:exp:value:FPC} includes values for continuous cost control, for variants of $\methodname$ using different $\Ucal$ (and should be viewed as an ablation study).
Note also how \baselineFullSuffix tends to perform badly despite searching through a much larger $\Ucal$. 
Like noted earlier, this is due to larger gap between $C(\hat{S})$ and $C^+$ as $|\Ucal|$ increases. 

On the cost side, \cref{table:main:exp:excesscost:FP} shows the mean \textbf{Excess Cost}, defined by the mean cost minus target cost (averaged across all targets).
\cref{table:main:exp:violation:FP} shows the mean \textbf{Violation Frequency}, namely the proportion of prediction sets with cost higher than target.
\baselineClasswise is conservative, verifying~\cref{sec:method:prelimsplitcp}, and \baselineInnerSet is too as it controls the probability that \textit{any} class is wrong.
Like \methodname, \baselineFPCP controls $C_{FP}$ well (but achieves lower values and cannot control general costs).
Results on continuous costs are in the Appendix.

Finally, results for \baselineFPCP are computed using \datastructname.
As shown in \cref{fig:main:exp:time}, \textbf{Computation Time} is significantly improved by \datastructname.
Even with binary search on the thresholds, the original method still takes too long to finish.


\section{Conclusion}
We have formalized a new problem of value-maximization with cost control for multi-label prediction sets, and proposed a framework, \methodname, to solve this problem. 
Improving upon previous works, \methodname explicitly maximizes a value function for the prediction set, and is able to control, in either expectation or violation risk, any continuous cost function. 
Moreover, with the help of \datastructname, a new custom data structure, \methodname can perform online updates efficiently for large-scale applications. 
Empirical evidence suggests that \methodname tends to achieve higher value while providing rigorous cost control. 
Since the trade-off between value and cost is common in many areas,  we hope this paper will provide useful insights for various application areas, as well as serve as a foundation for potential future research in this task.
For example, examining better $\Ucal$ as well as different $\hat{C}$ and $\hat{V}$ for specific value and cost functions.

\bibliography{main}
\bibliographystyle{icml2023}
\newpage
\appendix
\onecolumn
\section{Proofs}\label{appendix:sec:proofs}
\subsection{Proof for \cref{thm:splitcp:binary}}
This is a relatively standard result and can be found, for instance, in~\cite{Lei2014ClassificationConfidence}.
Hence we instead present a brief justification.

\textit{Proof}:
First of all, since $[Z_1, \ldots, Z_{N+1}]$ are assumed to be i.i.d., if we focus on the subset for which $Y_k=0$, they are still i.i.d. (with the new distribution being the conditional distribution of $(X|_{Y_k=0}, Y|_{Y_k=0})$), thus exchangeable.
If $Y_{N+1,k}=1$, then the statement we want to prove is vacuously true. 
For simplicity, we denote the total number of samples with $Y_k=0$ still as $N+1$, and will now assume all data have $Y_k=0$ (as we have restricted to this distribution).
Now, we could invoke the split conformal procedure from \ref{thm:splitcp:general}, and arrive at:
\begin{align}
    \mathbb{P}\{p_k(X_{N+1}) \leq Q(1-\epsilon; \{p_k(X_{i})\} \cup \{\infty\}\} &\geq 1-\epsilon\\
    \implies \mathbb{P}\{p_k(X_{N+1}) \leq t\} &\geq 1-\epsilon\\
    \implies \mathbb{P}\{k\in S_{N+1}\} &< \epsilon\\
\end{align}

Note that if we are to assume exchangeability, we need to assume exchangeability of all conditional distributions (with different $Y$), making it not too different from just assuming the stronger assumption of i.i.d. 
\qed

\subsection{Proof for \cref{thm:splitcp:classwise}}
\textit{Proof}:
\begin{align}
    \mathbb{E}[C_{FP}(S_{N+1}, Y_{N+1})] &= \mathbb{E}[|\{k: k\in S_{N+1}, k\not\in Y_{N+1}\}|]\\
     &= \mathbb{E}[\sum_{k\in [K]} \mathbbm{1}\{k: k\in S_{N+1}, k\not\in Y_{N+1}\}]\\
     &= \sum_{k\in [K]} \mathbb{P}\{k: k\in S_{N+1}, k\not\in Y_{N+1}\}\\
     &= \sum_{k\in [K]} \mathbb{P}\{k\not\in Y_{N+1}\}\mathbb{P}\{k: k\in S_{N+1}| k\not\in Y_{N+1}\}\\
     &\leq \sum_{k\in [K]} \mathbb{P}\{k: k\in S_{N+1}| k\not\in Y_{N+1}\} \label{eq:appendix:proof:eachclass}
\end{align}
Applying \cref{thm:splitcp:binary} to each $k$ for \cref{eq:appendix:proof:eachclass} (with $\epsilon=\frac{c}{K}$), we have
\begin{align}
    \mathbb{E}[C_{FP}(S_{N+1}, Y_{N+1})] &\leq \sum_{k\in[K]} \frac{c}{K} = c.
\end{align}
\qed

\newpage
\subsection{Proof for \cref{thm:control:expectation}}
Like~\cite{reffp}, the proof below uses a result from~\cite{riskcontrolproof}.
We modified the proof in \cite{riskcontrolproof} so it is simpler and more self-contained. 

\textit{Proof}:
Denote $C^+_i(\cdot)=C^+_i(\cdot,Z_i)$ for convenience.
$[C^+_1,\ldots,C^+_{N+1}]$ are thus exchangeable non-decreasing bounded functions in $[0,C_{max}]$. 
Due to exchangeability, 
\begin{align}
    \forall i\in[N+1], \forall t, \mathbb{E}[C^+_i(t)] \equiv \frac{\sum_{i=1}^{N+1} \mathbb{E}[C^+_i(t)]}{N+1} = \frac{\mathbb{E}[\sum_{i=1}^{N+1}C^+_i(t)]}{N+1}. \label{eq:appendix:proof:exchangemeansum}
\end{align}

Let $c'=c-\frac{C_{max}}{N+1}$.
We rewrite the definition in \cref{eq:expected:threshold} for $T_c$ as well as a new quantify $T'_c$:
\begin{align}
    T_c = \sup\{t: \frac{1}{N+1}\sum_{i=1}^{N} C^+_i(t) \leq c'\}\\
    T'_c = \sup\{t: \frac{1}{N+1}\sum_{i=1}^{N+1} C^+_i(t) \leq c\}.
\end{align}
(Note that if we are taking $\sup$ over empty sets $T_c$ would be $-\infty$, and $\hat{S}=\empty$ which will trivially satisfy all cost requirements.)
Because 
\begin{align}
    \frac{1}{N+1}\sum_{i=1}^{N} C^+_i(t) \leq c-\frac{C_{max}}{N+1}\implies \frac{1}{N+1}\sum_{i=1}^{N+1} C^+_i(t) \leq c ,
\end{align}
we have $T'_c\geq T_c$, which means by definition of $T'_c$:
\begin{align}
    \frac{\sum_{i=1}^{N+1}C^+_i(T_c)}{N+1}\ &\leq c.
\end{align}
Invoking \cref{eq:appendix:proof:exchangemeansum} for $i=N+1$ we have
\begin{align}
    \mathbb{E}[C^+_{N+1}(T_c)] &\leq c.
\end{align}
Finally, by definition of $C^+$:
\begin{align}
    \mathbb{E}[C(\hat{S}_{N+1};Z_{N+1})] \leq \mathbb{E}[\max_{S\in \Ucal(X_{N+1})}\{C(S;Z_{N+1}): \hat{C}(S; X_{N+1}) < T_{c}] \leq c.
\end{align}
Note that all expectations are taken over the randomness of all of $[Z_1,\ldots,Z_{N+1}]$.
\qed

\newpage
\subsection{Proof for \cref{thm:control:violation}}
Although there is a similar result in~\cite{reffp} leveraging~\cite{riskcontrolproof}, we will use a \textit{different} proof approach here.
We will define $T_{c,\delta}$ in a seemingly different way that makes the guarantee self-evident.
Then, we will establish equivalence to \cref{eq:violation:threshold}.
Note we defined $T_{c,\delta}$ like in \cref{eq:violation:threshold} (which might look less familiar to readers versant in conformal prediction literature) in order to align with~\cite{reffp} as well as \cref{eq:expected:threshold} (and to make it clear why \datastructname can be deployed similarly in both versions of cost control).

\textit{Proof}:
Denote random variable 
\begin{align}
    T_i&\defeq \max_{t}\{t\in \mathcal{T}_i\cup\{\infty\} : C^+(t; Z_i) \leq c\} \\
    \text{where } \mathcal{T}_i &\defeq \{\hat{C}(S;X_i): S\in \Ucal(X_i)\}.
\end{align}
It is clear that $[T_1,\ldots,T_{N+1}]$ are exchangeable because $[Z_1,\ldots,Z_{N+1}]$ are exchangeable.
Note that $C^+(T_i;Z_i) \leq c$.
Observe crucially that
\begin{itemize}
    \item 
    Since $C^+(t;Z_i)$ is non-decreasing in $t$, we have $ C^+(t;Z_i) > c \geq  C^+(T_i;Z_i) \implies t > T_i$.
    \item 
    By the definition of $T_i$ we also note that $t > T_i \implies C^+(t;Z_i) > c$ (if $T_i=\infty$ this is vacuously true). 
\end{itemize}
Thus, we have $T_i=\sup\{t\in\mathbb{R}: C^+(t;Z_i) \leq c\}$, or equivalently:
\begin{align}
    t>T_i \Leftrightarrow C^+(t;Z_i) > c \label{eq:appendix:thres=cost}
\end{align}

If we choose $T_{c,\delta} = -Q(1-\delta;\{-T_i\}_{i=1}^N\cup\{\infty\})$, \cref{thm:splitcp:general} directly gives us:
\begin{align}
\mathbb{P}\{T_{N+1} \geq T_{c,\delta}\} = \mathbb{P}\{C^+(T_{c,\delta};Z_{N+1})\leq c\} \geq 1-\delta
\end{align}

Finally, like before:
\begin{align}
    \mathbb{P}\{C(\hat{S}_{N+1};Z_{N+1}) \leq c\} 
    &\geq \mathbb{P}\{\max_{S\in \Ucal(X)}\{C(S;Z): \hat{C}(S; X) < T_{c,\delta}\}\leq c\} \\
    &=\mathbb{P}\{C^+(T_{c,\delta};Z_{N+1})\leq c\} \geq 1-\delta
\end{align}

Now, we only need to show that the $T_{c,\delta}$ defined above is the same as the one in \cref{eq:violation:threshold}.
To see this:
\begin{align}
    -Q(1-\delta;\{-T_{i}\}_{i=1}^N\cup\{\infty\}) &= -\inf\{t: \sum_{i=1}^N \mathbbm{1}\{t\geq -T_i\} + \mathbbm{1}\{t\geq \infty\} \geq (1-\delta)(N+1)\}\\
    &= \sup\{t: \sum_{i=1}^N \mathbbm{1}\{t\leq T_i\} + \mathbbm{1}\{t\leq -\infty\} \geq (1-\delta)(N+1)\}\\
    &= \sup\{t: \sum_{i=1}^N \mathbbm{1}\{t\leq T_i\} \geq (1-\delta)(N+1)\}
\end{align}
Note again that $\mathbbm{1}\{t\leq T_i\} = \mathbbm{1}\{C^+(t;Z_i) \leq c\}$, which means
\begin{align}
    T_{c,\delta} &= \sup\{t: \sum_{i=1}^N \mathbbm{1}\{C^+(t;Z_i) \leq c\} \geq (1-\delta)(N+1)\}
\end{align}

\qed

\newpage
\subsection{Proof for \cref{thm:equivalence:quantile}}
\textit{Proof}:
For simplicity, write $M = \frac{c(N+1)-C_{max}}{W}$, and define:
\begin{align}
    T'_c &= \inf \{t:\sum_{i=1}^N \sum_{j} w_{i,(j)}\Delta_{\hat{c}_{i,(j)}} \geq M \}\\
    &=\inf \{t:\sum_{i=1}^N \sum_{j>1} \frac{(c^+_{i,(j)} - c^+_{i,(j-1)})}{W}\mathbbm{1}\{t \geq \hat{c}_{i,(j)}\} \geq M \}
\end{align}
Note that $c^+_{i,(0)}= c_{i,(0)} = \hat{c}_{i,(0)} = 0$ for any $i$, which means
\begin{align}
    \sum_{j>1} \frac{(c^+_{i,(j)} - c^+_{i,(j-1)})}{W}\mathbbm{1}\{t \geq \hat{c}_{i,(j)}\} &= \frac{1}{W} \max\{c^+_{i,(j)}: \hat{c}_{i,(j)}\leq t\}.
\end{align}
Thus, 
\begin{align}
    T'_c &= \inf\{t: \sum_{i=1}^N \max\{c^+_{i,(j)}: \hat{c}_{i,(j)}\leq t\} \geq WM\} \label{eq:appendix:proof:computation:infversion}
\end{align}
Note the similarity between \cref{eq:appendix:proof:computation:infversion} and what we want below (i.e. \cref{eq:expected:threshold}):
\begin{align}
    T_c &= \sup\{t: \sum_{i=1}^N C^+(t; Z_i) \leq M\} = \sup\{t: \sum_{i=1}^N \max\{c^+_{i,(j)}: \hat{c}_{i,(j)} < t\} \leq WM\} \label{eq:appendix:proof:computation:supversion}
\end{align}

Rewrite the empirical weighted CDF as 
\begin{align}
    F = \sum_{i=1}^N \sum_{j} w_{i,(j)}\Delta_{\hat{c}_{i,(j)}} = \sum_{l=1}^m w'_l \Delta_{t_l}
\end{align}
where $t_1<t_2<\ldots t_m$ are the distinct values in $\{\hat{c}_{i,j}\}$, and $w'_l = \sum_{(i,j): \hat{c}_{i,(j)}=t_l} w_{i,(j)}$.
WLOG, assume $T'_c=t_{l^*}$.
Also write a version that is not ``inclusive'' of the threshold, defined as\begin{align}
    F'(t) = \sum_{i=1}^N \sum_{j} w_{i,(j)}\mathbbm{1}\{t > \hat{c}_{i,(j)}\} = \sum_{l=1}^m w'_l\mathbbm{1}\{t > t_l\}.
\end{align}
Note the sign in the indicator function is now $>$ instead of $\geq$. 
It is now clear that \cref{eq:appendix:proof:computation:infversion} and \cref{eq:appendix:proof:computation:supversion}  can be written as:
\begin{align}
    T'_c &= \inf\{t: F(t) \geq WM\} \label{eq:appendix:proof:computation:infversion:short}\\
    T_c &= \sup\{t: F'(t) \leq WM\} \label{eq:appendix:proof:computation:supversion:short}
\end{align}
Consider $t_{l^*-1}$, the immediate predecessor of $T'_c$.
By definition of $T'_c$ (\cref{eq:appendix:proof:computation:infversion:short}) we know 
\begin{align}
    F(t_{l^*-1}) < WM.
\end{align}
Moreover, by definition of $F$ and $F'$ we have
\begin{align}
    F(T'_c) - w'_{l^*} = F'(T'_c)  = F(t_{l^*-1}) < WM.
\end{align}
Following a similar logic, we know that $F'(t_{l^*+1}) = F(T'_c)$.
\begin{itemize}
    \item If $F(T'_c) = WM$ exactly, then technically $T_c$ is $t_{l^*+1}$. 
    However, the probability of this happening is very low (if $\hat{C}$ is continuous, such as those constructed from $\phat$).

    \item Otherwise, $F(T'_c) > WM$, which means $T_c = T'_c$.
\end{itemize}
\qed

\newpage
\section{Details on the Quantile Tracking Algorithms}\label{appendix:sec:algodetail}
\subsection{Maintenance of \datastructname}
Our exact implementation can be found in the supplementary material. 
The base structure of \datastructname is a red-black tree~\cite{guibas1978dichromatic}. 
Many implementations could be found online, and we based ours on \url{https://github.com/Bibeknam/algorithmtutorprograms}.
Like shown in \cref{alg:main:tree}, we add two fields to a \texttt{Node}: sum and weight.
Since each node represents a point mass of the empirical CDF (corresponding to a value $v$), the weight is naturally the mass of this value.
In other words, a \datastructname with $\{\texttt{Node}(value=v_i, weight=w_i)\}$ represents the empirical CDF $\sum_{i}w_i\Delta_{v_i}$. 
(We use $\Delta_v$ instead of the more commonly used notation $\delta_v$ because $\delta$ is already used in \cref{problem:violation}). 
Finally, we maintain a segmentation tree using \texttt{Node.sum}, which keeps track of the sum of weights of the sub-tree rooted at this node. 

\subsection{Full Algorithm with \texttt{DELETE}}
Due to space constraint, we did not include the \texttt{DELETE} operation in \cref{alg:main:tree}.
In \cref{alg:appendix:tree:full}, we present the full algorithm with this operation. 
\cref{alg:appendix:expected:full} explains the case for expected cost control. 

\def \AlgAppendixQuantileTreeFull{
\begin{algorithm}[h]
\SetAlgoLined
\DontPrintSemicolon
\SetKwProg{Struct}{Struct}{:}{}
\Struct{Node}{
\texttt{color} \# For red/black marking\\
\texttt{value} \# Recording $v_i$ (the threshold in $\Delta_{v_i}$)\\
\texttt{sum} \# The sum of all $w_i$ under this node\\
\texttt{weight} \# $w_i$. \texttt{weight} (thus \texttt{sum}) is un-normalized.
}
\SetKwFunction{Finsert}{INSERT}
\SetKwProg{Fn}{Function}{:}{}
\Fn{\Finsert{value $v$, weight $w$}}{
    Insert a leaf Node(\texttt{value}=$c$,\texttt{weight}=$w$)\\
   Fix the tree to a valid red-black tree, while maintaining valid \texttt{sum} values.
}
\textbf{End Function}

\SetKwFunction{Fdelete}{DELETE}
\Fn{\Fdelete{value $v$, weight $w$}}{
    Find node for $v$, denoted as $node(v)$\\
    {\eIf{$ node(v).weight = w $}
            {Remove $node(v)$\\
            Fix the tree to a valid red-black tree, while maintaining valid \texttt{sum} values}
            {$ node(v).weight \gets node(v).weight-w$}
    }
}
\textbf{End Function}

\SetKwFunction{Fquery}{QUERY}
\Fn{\Fquery{quantile $q$}}{
    Compute the cumulative weight $cw=$\texttt{root.sum}$*q$.\\
    Find the node($cw$) corresponding to $cw$ using binary search (via \texttt{sum}).\\
    \textbf{return} $node(cw).value$
}
\textbf{End Function}
\caption{\datastructname}
\label{alg:appendix:tree:full}
\end{algorithm}
}
\AlgAppendixQuantileTreeFull

\def \AlgAppendixExpectedCostFull{
\begin{algorithm}[h]
\SetAlgoLined
\DontPrintSemicolon
\texttt{tree} $\gets$ a \datastructname object\\
\SetKwFunction{Finsert}{INSERT}
\SetKwProg{Fn}{Function}{:}{}
\Fn{\Finsert{input $x_i$, label $y_i$}}{
    Denote $[\hat{c}_{i,(j)}]_{j=1}^M$ and $[w_{i,(j)}]_{j=1}^M$ as in \cref{thm:equivalence:quantile}. \\
    \ForEach{$ j = 1,\dots,M$}{
        \texttt{tree.INSERT}($\hat{c}_{i,(j)}, w_{i,(j)}$)
    }
    Cache $[c_{i,(j)}]_{j=1}^M$ and $\{w_{i,j}\}_{j=1}^M$ (for \texttt{DELETION}).
}
\textbf{End Function}

\SetKwFunction{Fdelete}{DELETE}
\Fn{\Fdelete{input $x_i$}}{
    Retrieve  $[\hat{c}_{i,(j)}]_{j=1}^M$ and $[w_{i,(j)}]_{j=1}^M$.\\
    \ForEach{$ j = 1,\dots,M$}{
        \texttt{tree.DELETE}($c_{i,(j)}, w_{i,(j)}$)
    }
}
\textbf{End Function}

\SetKwFunction{Fquery}{COMPUTE\_THRESHOLD}
\Fn{\Fquery{target cost $c$}}{
    \textbf{return} \texttt{tree.QUERY}($\frac{(N+1)\cdot c-C_{max}}{tree.root.sum}$)
}
\textbf{End Function}
\caption{Expected Cost Control}
\label{alg:appendix:expected:full}
\end{algorithm}
}
\AlgAppendixExpectedCostFull

\subsubsection{Maintaining \texttt{Node.sum}}
One technical detail we did not explain in \cref{alg:main:tree} is how to maintain valid \texttt{sum} while using the red-black tree. 
\begin{itemize}
    \item During rotations, we only need to fix the recursive relation of \texttt{Node.sum = Node.left.sum + Node.weight + Node.right.sum} in each rotation operation, by computing the new sums and replace the \texttt{sum} after the rotation. 
    For example, in a left rotation like:
    \begin{verbatim}
          x              y
           \      ->    /
            y          x
    \end{verbatim}
    We need to compute 
    \begin{verbatim}
    //BEGIN OF CODE======================================================
        new_x_sum = x.left.sum + y.left.sum + x.weight
        new_y_sum = new_x_sum + y.weight + y.right.sum
        ... //perform the rotation
        x.sum = new_x_sum
        y.sum = new_y_sum
    //END OF CODE========================================================
    \end{verbatim}

    \item Other than rotations, we only need to update the \texttt{sum} of all $O(log(N))$ nodes along the path to the node being inserted/deleted/updated.
\end{itemize}

\subsection{Online Computation for $T_{c,\delta}$}\label{appendix:sec:algodetail:violation}
In \cref{alg:appendix:violation}, we present the computation of $T_{c,\delta}$ in an online fashion.
Note that for this case, since it only involves counting integers (i.e. the weight is $1$ for all thresholds), we could use existing python packages like \texttt{sortedcontainers.SortedList} to replace the self-balancing segmentation binary search tree.
\def \AlgAppendixViolation{
\begin{algorithm}[h]
\SetAlgoLined
\DontPrintSemicolon
\texttt{tree} $\gets$ a self-balancing segmentation binary search tree\\
Fix the cost target $c$\\
\SetKwFunction{Finsert}{INSERT}
\SetKwProg{Fn}{Function}{:}{}
\Fn{\Finsert{input $x_i$, label $y_i$}}{
    Let $t_i\gets \inf_t\{C^+(t;(x_i,y_i) > c\}$\\
    \texttt{tree.INSERT}($t_i, weight=1$)\\
    Cache $t_i$ (for \texttt{DELETION}).
}
\textbf{End Function}

\SetKwFunction{Fdelete}{DELETE}
\Fn{\Fdelete{input $x_i$}}{
    Retrieve  $t_i$.\\
    \texttt{tree.DELETE}($t_i, weight=1$)
}
\textbf{End Function}

\SetKwFunction{Fquery}{COMPUTE\_THRESHOLD}
\Fn{\Fquery{violation goal $\delta$}}{
    Compute quantile to query $q=\frac{(N+1)\delta-1}{tree.root.sum}$.\\
    \textbf{return} \texttt{tree.QUERY}($q$) 
}
\textbf{End Function}
\caption{Violation Control}
\label{alg:appendix:violation}
\end{algorithm}
}
\AlgAppendixViolation

\subsection{An Example Application of \datastructname to Existing Conformal Prediction Methods}
In \cref{alg:main:expected}, we saw how \datastructname could be used to efficiently update the empirical CDF and look up $Q(\beta;F)$ for any $\beta$. 
We now look at a slightly different potential usage of \datastructname in a recent paper~\cite{bates-rcps}, by efficiently evaluating $F$.
(Note that $F=Q^{-1}$ for an continuous distribution.)
We try to summarize the important steps in~\cite{bates-rcps} here, but readers should refer to the original paper for a better understanding of their method.
Like noted before, RCPS~\cite{bates-rcps} controls the violation probability of the population risk.
It achieves so by using concentration bounds to upper-bound the population risk, which is a function of a threshold parameter $\lambda$.
The final prediction set is given as $\hat{S}(x) = \{k: \hat{p}_k(x) > \hat{\lambda}\}$, where $\hat{\lambda}$ is the chosen threshold.

In practice, $\lambda$ could be searched from all applicable thresholds on the calibration set, and the bound is computed using the calibration set as well.
Assuming the more tractable Hoeffding-Bentkus version is being used, to compute this upper-bound (for any $\lambda$), we need to compute the empirical risk $\hat{R}(\lambda)$.
In~\cite{bates-rcps}, $\hat{R}(\lambda)$ only needs to be evaluated once, on the entire calibration set.
As a result, one could cache the results $\hat{R}(\lambda)$ for each $\lambda$ ($O(NK)$) and then perform a binary search over $\lambda$ ($O(\log{(NK)})$), like what was done in the code published on \url{https://github.com/aangelopoulos/rcps}.
Assuming we now want to update the threshold $\hat{\lambda}$ in an online fashion, which requires us to update $\hat{R}(\lambda)$ for \textit{every possible} $\lambda$ whenever the $N$-th calibration data point comes in, for $N=1, 2,\ldots$.  
Actually recomputing $\hat{R}(\lambda)$ is obviously inefficient, but even if we only compute $R_{N+1}(\lambda)$ for each $\lambda_1,\ldots,\lambda_{NK}$ ($O(K^2N)$), and $R_{i}(\lambda_j)$ for $i\in[N]$ and $j\in [K$] ($O(K^2N)$), it is still an $O(K^2N + NK\log{(NK)})$ operation per update, or $O(N\log{N})$ assuming $N\gg K$ for simplicity).
If we further use data structures like \texttt{sortedcontainers.SortedList} to maintain the sorted list of all $\{\lambda_{i,k}\}_{i\in[N],k\in[K]}$, the complexity for each update+query can reduce to $O(K^2N + \log{(NK)})$ (or $O(N)$ when $N\gg K$).

Instead, we could leverage the fact that $\hat{R}$ is monotonic decreasing in $\lambda$, and store the following CDF in a \datastructname:
\begin{align}
    F &= \sum_{i=1}^N\sum_{j=1}^K w_{i,(j)} \Delta_{-l_{i,(j)}}\label{eq:appendix:rcpscdf}\\
    \text{ where } w_{i,(j)} &= R_i(l_{i,(j)}) - R_i(l_{i,(j+1)})
\end{align}
Here, we suppose the prediction (logits) are already sorted in a way that $j\leq j'\implies l_{i,(j)} \leq l_{i,(j')}$ (takes $O(K\log{K})$), and that the risk of an empty set is zero\footnote{If not, we just need to adjust the loss on a $\emptyset$ from all our estimate of $\hat{R}(\lambda)$.} (i.e. $R_i(l_{i,K})=0$).
Now, we could easily query $\hat{R}(\lambda) = F(-\lambda)$ in $O(\log{NK})$ time, reducing the total time per update+query to $(K^2 + K\log{K} + (\log{(NK)})^2)$, or $O((\log{N})^2)$ if $N\gg K$.
An implementation of an online version of RCPS is also included in the supplementary material.
We would like to emphasize that while \datastructname is given, identifying the correct CDF to query (like in \cref{eq:appendix:rcpscdf}) is crucial.

In fact, one can potentially improve the update time by first computing the max acceptable $\hat{R}$ given $N$ and $\delta$, denoted as $\hat{R}_{max}$,  and then directly query $Q(\hat{R}_{max};F)$, resulting in $O( K^2 + K\log{K} + \log{\frac{B}{b}} + \log{(NK)})$ time in total.
Here $O(\log{\frac{B}{b}})$ is the complexity of finding the $\hat{R}_{max}$ by plugging in different risk values into the Hoeffding-Bentkus bound with a (continuous) binary search, with $B$ denoting the width of the search interval and $b$ denoting the tolerance.
$B$ is always less than or equal to the target risk, and $b$ could be set to the minimum weight of the \datastructname (which takes $O(1)$ to track per update). 
This version has a more similar flavor as \cref{alg:main:expected}, in that it only query \datastructname once per update.
Depending on the specific distribution of interest, we might also see improved with this approach.

\newpage
\section{Additional Experiment Details}\label{appendix:sec:additionalexp}
In this section, we include additional experiment details, including results that cannot fit into the main paper. 

\subsection{Datasets}
\begin{itemize}
    \item \dataMNIST:
    We first split the training set of the original MNIST~\cite{mnist} into 90/10 train/validation split. 
    We then create 54000 RGB images of shape $3\times 48\times 48$. 
    For each image, we sample the 10 classes independently with probability 0.4.
    For each class, if it is to appear in the new image, we sample one original MNIST image from the training set and randomly assign it to a location and channel (the original MNIST is $1\times 28 \times 28$).
    The 10000 new test images are generated the same way but with original images from the test set only.
    
    \item \dataMIMICThree~\cite{MIMIC,PhysioNet,MIMICDemo}:
    We select all admissions with an non-empty discharge report to simulate the task to predict missing diagnoses.
    For each admission, aggregate all ICD-9-CM diagnosis codes associated with the admission, map them to hierarchical condition category (HCC) code, and randomly delete the codes with a probability of 30\%.
    We use the 3-digit version of ICD codes, and drop the bottom rare codes accounting for 3\% of the total frequency. 
    The deleted codes are used as target of prediction, and the un-deleted codes are taken as input to the model.
    Other input features used include patient features pre-processed like in~\cite{mimic3preprocessing}, as well as clinical notes prepared using~\cite{clinicalbert}.
    We delete the typical section in clinical notes that list all diagnoses or disease name, so they do not give away the answer directly.
    We restricted the subset of HCC codes to the ones with at least 1000 patients, which is the same as using a code frequency cutoff of 0.8\%.
    The precise list of HCC codes could be found in the code in our supplementary material.  
    The risk factors are taken from \url{https://www.cms.gov/medicare/health-plans/medicareadvtgspecratestats/downloads/announcement2014.pdf}. 
    Note that we present $V_{TPC}$ and $C_{FPC}$ normalized to $[0,100]$.
    One could recover the total risk adjustment factor by using the $C_{max}$, which is 6.56 for \dataMIMICThree(select) and 16.48 for \dataMIMICThree.

    \item  \dataClaim: Insurance claim prediction on a proprietary dataset from a large healthcare data provider in North America.
    We kept only ICD-10-CM records (from 2015-10-01). 
    Each claim will be treated as one sample.
    The input is all the ICD-10-CM codes in the past 26 weeks, and the target of prediction are the codes for this patient in the next 26 weeks. 
    Like in \dataMIMICThree, we set the code frequency threshold at 0.8\%. 
    The precise list of HCC codes could be found in the code in our supplementary material.  
    The risk factors are taken from \url{https://www.cms.gov/medicare/health-plans/medicareadvtgspecratestats/risk-adjustors/2023-model-software/icd-10-mappings} (2023 Midyear Software Model).
    For \dataClaimSeq, we used a temporal split of 75/25 (or the cutoff date of 2017-09-03) for training vs test set, resulting in 233679 samples for training and 77894 samples for testing.
    For \dataClaim, the training set and test contains random but mutuallly exclusive patients.
    The total risk adjustment factor can be recovered by using $C_{max}$ of 7.41 for \dataClaim(select) and 14.87 for \dataClaim.
\end{itemize}

\subsection{Base Neural Networks}
For each task, all methods presented in the paper are applied on the same base classifiers $\phat$, which are neural networks further described below:
\begin{itemize}
    \item \dataMNIST:
    We used a Convolutional Neural Network consisting of three \texttt{Conv-BatchNorm-ReLU-MaxPool} blocks~\cite{BatchNorm,lecun2015deep}, with a fully connected layer at the end.
    We train the model with ADAM~\cite{kingma2014adam} with a learning rate of 1e-2 and batch size of 128 for 50 epochs.

    \item \dataMIMICThree:
    We used a ClinicalBERT~\cite{clinicalbert} to encode the clinical notes.
    Features processed by~\cite{mimic3preprocessing} (including some biomarkers) are fed into a multi-layer perceptron.
    Finally, embeddings (as well as HCC codes that are not deleted) are concatenated and fed into the final prediction layers.
    We train the model with ADAM~\cite{kingma2014adam} with a learning rate of 1e-5 and batch size of 16 for 50 epochs.

    \item \dataClaim (and \dataClaimSeq):
    We used a model with two layers of self-attention~\cite{vaswani2017attention} to aggregate code embeddings to visit embeddings, as well as visit embeddings to an overall embedding, which is then concatenated with demographic information before being fed into the final prediction layer. 
    We train the model with ADAM~\cite{kingma2014adam} with a learning rate of 1e-5 and batch size of 128 for 20 epochs.

\end{itemize}
For $\baselineFPCP$ we need to train a DeepSets-based cost proxy, so we will further perform a 80/20 split on the training data, using 80\% to train the base classifier and 20\% to train the cost proxy.
The exact training scripts and model definitions could be found in our code. All experiments are carried out with PyTorch~\cite{PyTorch}.

\subsection{Additional Results}
We include completed experiment results in \cref{table:appendix:exp:value:fp} (Value for $C_{FP}$ control), \cref{table:appendix:exp:value:cts} (Value for $C_{FPC}$ control), \cref{table:appendix:exp:cost:fp} (Excess Cost and Violation for $C_{FP}$ control) and \cref{table:appendix:exp:cost:cts} (Excess Cost and Violation for $C_{FPC}$ control).
Methods in \cref{table:appendix:exp:value:cts} and \cref{table:appendix:exp:cost:cts} could be considered ablation study of \methodname with different $\Ucal$.
Note that the ``(NN)'' suffix to method names means using the DeepSets-based cost proxy like \baselineFPCP to replace the $\hat{C}$ introduced in \cref{sec:method:proxy}.
The conclusions are essentially the same as those presented in the main paper.

\def \TableAppendixFPValueFull{
\begin{table*}[ht]
\caption{
Mean value of the prediction sets with false positive control (expectation control as well as violation control).
The higher the better. 
}
\label{table:appendix:exp:value:fp}
\vskip 0.15in
\begin{center}
\begin{small}
   \resizebox{1\columnwidth}{!}{
\begin{tabular}{l|cc|cccccc|c}
\toprule
  & & & \multicolumn{7}{|c}{Variants of \methodname}\\
Expectation Control
  & \baselineClasswise & \baselineInnerSet & \baselineFullSuffix & \baselineGreedyProbSuffix(NN) (\baselineFPCP) & \baselineGreedyValSuffix(NN) & \baselineGreedyRatioSuffix(NN) & \baselineGreedyProbSuffix & \baselineGreedyValSuffix & \baselineGreedyRatioSuffix\\
\midrule
\dataMIMICThree(select) & 2.137$\pm$0.081 & 0.869$\pm$0.064 & 2.186$\pm$0.074 & 2.159$\pm$0.071 & 2.241$\pm$0.078 & 2.245$\pm$0.081 & 2.188$\pm$0.066 & \textbf{2.263$\pm$0.078} & 2.261$\pm$0.078\\
\dataMIMICThree & 1.443$\pm$0.053 & 0.338$\pm$0.029 & \textendash & 1.564$\pm$0.044 & 1.607$\pm$0.047 & 1.610$\pm$0.047 & 1.587$\pm$0.046 & \textbf{1.636$\pm$0.049} & \textbf{1.637$\pm$0.049}\\
\dataClaim(select) & 1.829$\pm$0.113 & 1.446$\pm$0.065 & 2.061$\pm$0.109 & 2.022$\pm$0.102 & 2.094$\pm$0.114 & \textbf{2.106$\pm$0.116} & 2.057$\pm$0.101 & 2.108$\pm$0.113 & \textbf{2.113$\pm$0.112}\\
\dataClaim & 1.168$\pm$0.071 & 0.682$\pm$0.040 & \textendash & 1.179$\pm$0.067 & 1.219$\pm$0.067 & \textbf{1.223$\pm$0.068} & 1.191$\pm$0.067 & \textbf{1.218$\pm$0.064} & \textbf{1.221$\pm$0.065}\\
\dataClaimSeq(select) & 1.892 & 1.523 & 2.147 & 2.043 & 2.165 & 2.170 & 2.100 & 2.186 & \textbf{2.190}\\
\dataClaimSeq & 1.224 & 0.747 & \textendash & 1.209 & 1.264 & 1.266 & 1.242 & 1.285 & \textbf{1.286}\\
\dataMNIST & 35.342$\pm$0.231 & 24.058$\pm$0.339 & 36.570$\pm$0.233 & 37.442$\pm$0.274 & 37.893$\pm$0.284 & 37.953$\pm$0.280 & 37.714$\pm$0.245 & 38.060$\pm$0.260 & \textbf{38.203$\pm$0.259}\\
\dataMNIST(GEN) & 35.874$\pm$0.280 & 24.764$\pm$0.353 & 38.153$\pm$0.315 & 38.312$\pm$0.323 & 39.415$\pm$0.335 & 39.557$\pm$0.325 & 38.552$\pm$0.315 & 39.478$\pm$0.346 & \textbf{39.637$\pm$0.337}\\

\midrule
Violation Control\\
\midrule
\midrule
\dataMIMICThree(select) & \textendash & 0.304$\pm$0.039 & 2.188$\pm$0.073 & 2.080$\pm$0.072 & 2.165$\pm$0.076 & 2.174$\pm$0.082 & 2.116$\pm$0.063 & \textbf{2.211$\pm$0.074} & 2.206$\pm$0.075\\
\dataMIMICThree & \textendash & 0.137$\pm$0.022 & \textendash & 1.411$\pm$0.043 & 1.480$\pm$0.046 & 1.467$\pm$0.043 & 1.559$\pm$0.047 & 1.610$\pm$0.048 & \textbf{1.611$\pm$0.048}\\
\dataClaim(select) & \textendash & 1.016$\pm$0.049 & 2.067$\pm$0.110 & 1.997$\pm$0.098 & 2.071$\pm$0.114 & \textbf{2.082$\pm$0.118} & 2.027$\pm$0.096 & 2.081$\pm$0.110 & \textbf{2.085$\pm$0.114}\\
\dataClaim & \textendash & 0.420$\pm$0.020 & \textendash & 0.910$\pm$0.052 & 0.955$\pm$0.054 & 0.963$\pm$0.053 & 1.172$\pm$0.066 & 1.203$\pm$0.063 & \textbf{1.206$\pm$0.063}\\
\dataClaimSeq(select) & \textendash & 1.089 & 2.152 & 2.025 & 2.147 & 2.150 & 2.072 & 2.163 & \textbf{2.167}\\
\dataClaimSeq & \textendash & 0.466 & \textendash & 0.964 & 1.021 & 1.022 & 1.226 & 1.271 & \textbf{1.272}\\
\dataMNIST & \textendash & 16.120$\pm$0.538 & 35.262$\pm$0.212 & 36.390$\pm$0.264 & 36.926$\pm$0.293 & 36.974$\pm$0.304 & 36.695$\pm$0.219 & 37.131$\pm$0.240 & \textbf{37.310$\pm$0.249}\\
\dataMNIST(GEN) & \textendash & 16.836$\pm$0.573 & 36.886$\pm$0.298 & 37.256$\pm$0.303 & 38.488$\pm$0.322 & 38.630$\pm$0.341 & 37.509$\pm$0.295 & 38.614$\pm$0.335 & \textbf{38.804$\pm$0.338}\\
\bottomrule
\end{tabular}
}
\end{small}
\end{center}
\vskip -0.1in
\end{table*}
}
\TableAppendixFPValueFull

\def \TableAppendixCTSValueFull{
\begin{table}[ht]
\caption{
Mean value of the prediction sets with continuous cost ($C_{FPC}$) control (expectation control as well as violation control).
The higher the better. 
The difference is only the $\Ucal$ being used. 
}
\label{table:appendix:exp:value:cts}
\vskip -0.15in
\begin{center}
\begin{small}
   \resizebox{1\columnwidth}{!}{
\begin{tabular}{l|cccc|cccc}
\toprule
 & \multicolumn{4}{|c}{Expectation Control} & \multicolumn{4}{|c}{Violation Control}\\
  & \baselineFullSuffix & \baselineGreedyProbSuffix & \baselineGreedyValSuffix & \baselineGreedyRatioSuffix & \baselineFullSuffix & \baselineGreedyProbSuffix & \baselineGreedyValSuffix & \baselineGreedyRatioSuffix\\
\midrule
\dataMIMICThree(select,TP) & 3.492$\pm$0.102 & \textbf{3.628$\pm$0.104} & \textbf{3.628$\pm$0.104} & \textbf{3.628$\pm$0.109} & 3.398$\pm$0.097 & 3.322$\pm$0.094 & 3.322$\pm$0.094 & \textbf{3.427$\pm$0.102}\\
\dataMIMICThree(TP) & \textendash & 1.914$\pm$0.025 & 1.914$\pm$0.025 & \textbf{1.948$\pm$0.026} & \textendash & 1.837$\pm$0.027 & 1.837$\pm$0.027 & \textbf{1.891$\pm$0.025}\\
\dataClaim(select,TP) & 4.506$\pm$0.162 & 4.515$\pm$0.160 & 4.515$\pm$0.160 & \textbf{4.555$\pm$0.163} & 4.424$\pm$0.157 & 4.438$\pm$0.159 & 4.438$\pm$0.159 & \textbf{4.451$\pm$0.161}\\
\dataClaim(TP) & \textendash & 1.849$\pm$0.082 & 1.849$\pm$0.082 & \textbf{1.878$\pm$0.086} & \textendash & 1.808$\pm$0.080 & 1.808$\pm$0.080 & \textbf{1.839$\pm$0.083}\\
\dataClaimSeq(select,TP) & 4.506 & 4.525 & 4.525 & \textbf{4.545} & 4.430 & \textbf{4.462} & \textbf{4.462} & 4.453\\
\dataClaimSeq(TP) & \textendash & 1.856 & 1.856 & \textbf{1.877} & \textendash & 1.825 & 1.825 & \textbf{1.847}\\
\dataMNIST(TP) & 34.625$\pm$0.263 & 36.801$\pm$0.287 & 36.801$\pm$0.287 & \textbf{37.046$\pm$0.289} & 31.103$\pm$0.225 & 32.760$\pm$0.264 & 32.760$\pm$0.264 & \textbf{33.485$\pm$0.254}\\
\midrule
\dataMIMICThree(select,TPC) & 1.935$\pm$0.048 & 2.089$\pm$0.053 & \textbf{2.122$\pm$0.062} & 2.089$\pm$0.053 & \textbf{1.855$\pm$0.046} & \textbf{1.856$\pm$0.046} & 1.721$\pm$0.048 & \textbf{1.856$\pm$0.046}\\
\dataMIMICThree(TPC) & \textendash & \textbf{1.588$\pm$0.040} & 1.499$\pm$0.040 & \textbf{1.588$\pm$0.040} & \textendash & \textbf{1.513$\pm$0.041} & 1.376$\pm$0.036 & \textbf{1.513$\pm$0.041}\\
\dataClaim(select,TPC) & 2.045$\pm$0.084 & \textbf{2.080$\pm$0.087} & 2.019$\pm$0.097 & \textbf{2.080$\pm$0.087} & 1.994$\pm$0.080 & \textbf{2.020$\pm$0.082} & 1.879$\pm$0.078 & \textbf{2.020$\pm$0.082}\\
\dataClaim(TPC) & \textendash & \textbf{1.238$\pm$0.064} & 1.181$\pm$0.060 & \textbf{1.238$\pm$0.064} & \textendash & \textbf{1.206$\pm$0.062} & 1.141$\pm$0.058 & \textbf{1.206$\pm$0.062}\\
\dataClaimSeq(select,TPC) & 2.077 & \textbf{2.115} & 2.111 & \textbf{2.115} & 2.028 & \textbf{2.051} & 1.972 & \textbf{2.051}\\
\dataClaimSeq(TPC) & \textendash & \textbf{1.272} & 1.243 & \textbf{1.272} & \textendash & \textbf{1.245} & 1.204 & \textbf{1.245}\\
\dataMNIST(TPC) & 33.871$\pm$0.221 & \textbf{36.649$\pm$0.248} & 35.987$\pm$0.256 & \textbf{36.649$\pm$0.248} & 30.024$\pm$0.220 & \textbf{32.374$\pm$0.261} & 30.759$\pm$0.290 & \textbf{32.374$\pm$0.261}\\
\dataMNIST(GEN) & 36.542$\pm$0.272 & 37.598$\pm$0.312 & 38.386$\pm$0.341 & \textbf{38.870$\pm$0.321} & 33.170$\pm$0.269 & 33.458$\pm$0.302 & 34.398$\pm$0.358 & \textbf{35.645$\pm$0.335}\\
\bottomrule
\end{tabular}
}
\end{small}
\end{center}
\vskip -0.1in
\end{table}
}
\TableAppendixCTSValueFull

\def \TableAppendixFPCostFull{
\begin{table*}[ht]
\caption{
Excess cost and violation frequency for false positive control.
Values lower than 0 (for excess cost) or $\delta=0.1$ (for violation) are marked \costfail{red}.
Note we do not mark \dataClaimSeq as we cannot repeat the experiment. 
However, the conclusion is similar to \dataClaim.
}
\label{table:appendix:exp:cost:fp}
\vskip 0.15in
\begin{center}
\begin{small}
   \resizebox{1\columnwidth}{!}{
\begin{tabular}{l|cc|cccccc|c}
\toprule
  & & & \multicolumn{7}{|c}{Variants of \methodname}\\
Excess Cost
& \baselineClasswise & \baselineInnerSet & \baselineFullSuffix & \baselineGreedyProbSuffix(NN) (\baselineFPCP) & \baselineGreedyValSuffix(NN) & \baselineGreedyRatioSuffix(NN) & \baselineGreedyProbSuffix & \baselineGreedyValSuffix & \baselineGreedyRatioSuffix\\
\midrule
\dataMIMICThree(select) & \costfail{-2.32$\pm$0.46} & \costfail{-26.58$\pm$0.13} & \costfail{-2.51$\pm$0.06} & 0.02$\pm$0.15 & -0.01$\pm$0.15 & -0.00$\pm$0.15 & 0.00$\pm$0.16 & -0.02$\pm$0.16 & -0.01$\pm$0.15\\
\dataMIMICThree & \costfail{-0.91$\pm$0.36} & \costfail{-27.10$\pm$0.05} & \textendash & -0.04$\pm$0.08 & -0.04$\pm$0.07 & -0.04$\pm$0.07 & -0.05$\pm$0.06 & -0.05$\pm$0.05 & -0.05$\pm$0.05\\
\dataClaim(select) & \costfail{-2.48$\pm$0.51} & \costfail{-23.40$\pm$0.26} & \costfail{-2.42$\pm$0.09} & -0.03$\pm$0.19 & -0.03$\pm$0.17 & -0.02$\pm$0.18 & -0.04$\pm$0.20 & -0.03$\pm$0.19 & -0.04$\pm$0.20\\
\dataClaim & \costfail{-1.15$\pm$0.57} & \costfail{-24.40$\pm$0.19} & \textendash & -0.03$\pm$0.06 & -0.03$\pm$0.06 & -0.03$\pm$0.06 & -0.06$\pm$0.06 & -0.05$\pm$0.05 & -0.05$\pm$0.05\\
\dataClaimSeq(select) & -2.68 & -24.00 & -2.32 & 0.01 & 0.00 & 0.00 & 0.00 & -0.00 & -0.00\\
\dataClaimSeq & -1.21 & -24.61 & \textendash & 0.03 & 0.03 & 0.03 & 0.01 & 0.01 & 0.01\\
\dataMNIST & \costfail{-13.14$\pm$0.25} & \costfail{-27.09$\pm$0.11} & \costfail{-6.78$\pm$0.10} & -0.08$\pm$0.10 & -0.09$\pm$0.09 & -0.09$\pm$0.10 & -0.06$\pm$0.11 & -0.08$\pm$0.11 & -0.07$\pm$0.12\\
\dataMNIST(GEN) & \costfail{-13.14$\pm$0.25} & \costfail{-27.09$\pm$0.11} & \costfail{-6.76$\pm$0.11} & -0.08$\pm$0.10 & -0.10$\pm$0.13 & -0.10$\pm$0.12 & -0.06$\pm$0.11 & \costfail{-0.10$\pm$0.07} & -0.08$\pm$0.09\\
\midrule
Violation Frequency (in $\%$)\\
\midrule
\midrule
\dataMIMICThree(select) & \textendash & \costfail{0.13$\pm$0.02} & \costfail{7.32$\pm$0.37} & 10.05$\pm$0.49 & 10.07$\pm$0.44 & 10.01$\pm$0.44 & 10.21$\pm$0.40 & 10.08$\pm$0.27 & 10.13$\pm$0.32\\
\dataMIMICThree & \textendash & \costfail{0.01$\pm$0.00} & \textendash & 10.00$\pm$0.48 & 10.00$\pm$0.46 & 10.05$\pm$0.52 & 9.96$\pm$0.60 & 9.94$\pm$0.55 & 9.95$\pm$0.55\\
\dataClaim(select) & \textendash & \costfail{0.30$\pm$0.07} & \costfail{7.38$\pm$0.73} & 9.94$\pm$0.69 & 10.02$\pm$0.62 & 10.03$\pm$0.64 & 9.83$\pm$0.87 & 9.74$\pm$0.73 & 9.75$\pm$0.73\\
\dataClaim & \textendash & \costfail{0.06$\pm$0.02} & \textendash & 10.08$\pm$0.34 & 10.11$\pm$0.32 & 10.12$\pm$0.34 & 9.76$\pm$0.53 & 9.89$\pm$0.44 & 9.84$\pm$0.45\\
\dataClaimSeq(select) & \textendash & 0.30 & 7.32 & 9.97 & 10.01 & 9.97 & 9.91 & 10.01 & 9.96\\
\dataClaimSeq & \textendash & 0.07 & \textendash & 10.22 & 10.27 & 10.26 & 10.18 & 10.21 & 10.22\\
\dataMNIST & \textendash & \costfail{0.10$\pm$0.03} & \costfail{1.55$\pm$0.12} & 9.88$\pm$0.42 & 9.81$\pm$0.27 & 9.85$\pm$0.34 & 9.97$\pm$0.54 & 9.92$\pm$0.43 & 9.91$\pm$0.39\\
\dataMNIST(GEN) & \textendash & \costfail{0.10$\pm$0.03} & \costfail{1.57$\pm$0.14} & 9.88$\pm$0.42 & 9.73$\pm$0.33 & 9.83$\pm$0.40 & 9.97$\pm$0.54 & 9.97$\pm$0.52 & 10.08$\pm$0.48\\
\bottomrule
\end{tabular}
}
\end{small}
\end{center}
\vskip -0.1in
\end{table*}
}
\TableAppendixFPCostFull

\def \TableAppendixCTSCostFull{
\begin{table}[ht]
\caption{
Excess cost and violation frequency for continuous cost ($C_{FPC}$) control.
Values lower than 0 (for excess cost) or $\delta=0.1$ (for violation) are marked \costfail{red}.
Note we do not mark \dataClaimSeq as we cannot repeat the experiment. 
However, the conclusion is similar to \dataClaim.
}
\label{table:appendix:exp:cost:cts}
\vskip -0.15in
\begin{center}
\begin{small}
   \resizebox{1\columnwidth}{!}{
\begin{tabular}{l|cccc|cccc}
\toprule
 & \multicolumn{4}{|c}{Expectation Control} & \multicolumn{4}{|c}{Violation Control}\\
  & \baselineFullSuffix & \baselineGreedyProbSuffix & \baselineGreedyValSuffix & \baselineGreedyRatioSuffix & \baselineFullSuffix & \baselineGreedyProbSuffix & \baselineGreedyValSuffix & \baselineGreedyRatioSuffix\\
\midrule
\dataMIMICThree(select,TP) & \costfail{-3.49$\pm$0.05} & -0.00$\pm$0.12 & -0.00$\pm$0.12 & 0.00$\pm$0.08 & \costfail{4.35$\pm$0.26} & 10.04$\pm$0.42 & 10.04$\pm$0.42 & 10.08$\pm$0.35\\
\dataMIMICThree(TP) & \textendash & -0.04$\pm$0.07 & -0.04$\pm$0.07 & -0.06$\pm$0.06 & \textendash & 9.84$\pm$0.50 & 9.84$\pm$0.50 & 9.87$\pm$0.53\\
\dataClaim(select,TP) & \costfail{-3.14$\pm$0.06} & -0.05$\pm$0.08 & -0.05$\pm$0.08 & -0.05$\pm$0.10 & \costfail{5.65$\pm$0.44} & 9.92$\pm$0.59 & 9.92$\pm$0.59 & 9.77$\pm$0.62\\
\dataClaim(TP) & \textendash & -0.06$\pm$0.07 & -0.06$\pm$0.07 & -0.05$\pm$0.05 & \textendash & 9.71$\pm$0.44 & 9.71$\pm$0.44 & 9.70$\pm$0.41\\
\dataClaimSeq(select,TP) & -3.15 & 0.01 & 0.01 & 0.02 & 5.11 & 9.84 & 9.84 & 9.82\\
\dataClaimSeq(TP) & \textendash & 0.01 & 0.01 & 0.01 & \textendash & 9.99 & 9.99 & 10.02\\
\dataMNIST(TP) & \costfail{-7.46$\pm$0.10} & -0.08$\pm$0.09 & -0.08$\pm$0.09 & -0.08$\pm$0.07 & \costfail{3.46$\pm$0.11} & 9.90$\pm$0.38 & 9.90$\pm$0.38 & 9.92$\pm$0.32\\
\midrule
\dataMIMICThree(select,TPC) & \costfail{-2.71$\pm$0.05} & -0.00$\pm$0.12 & -0.06$\pm$0.17 & -0.00$\pm$0.12 & \costfail{5.56$\pm$0.31} & 10.04$\pm$0.42 & 9.93$\pm$0.30 & 10.04$\pm$0.42\\
\dataMIMICThree(TPC) & \textendash & -0.04$\pm$0.07 & -0.05$\pm$0.06 & -0.04$\pm$0.07 & \textendash & 9.84$\pm$0.50 & 9.91$\pm$0.56 & 9.84$\pm$0.50\\
\dataClaim(select,TPC) & \costfail{-2.35$\pm$0.07} & -0.05$\pm$0.08 & -0.09$\pm$0.13 & -0.05$\pm$0.08 & \costfail{7.31$\pm$0.56} & 9.92$\pm$0.59 & 9.97$\pm$0.31 & 9.92$\pm$0.59\\
\dataClaim(TPC) & \textendash & -0.06$\pm$0.07 & -0.04$\pm$0.05 & -0.06$\pm$0.07 & \textendash & 9.71$\pm$0.44 & 9.90$\pm$0.30 & 9.71$\pm$0.44\\
\dataClaimSeq(select,TPC) & -2.48 & 0.01 & -0.04 & 0.01 & 6.59 & 9.84 & 10.00 & 9.84\\
\dataClaimSeq(TPC) & \textendash & 0.01 & 0.03 & 0.01 & \textendash & 9.99 & 10.16 & 9.99\\
\dataMNIST(TPC) & \costfail{-6.94$\pm$0.09} & -0.08$\pm$0.09 & -0.11$\pm$0.11 & -0.08$\pm$0.09 & \costfail{3.98$\pm$0.14} & 9.90$\pm$0.38 & 9.82$\pm$0.32 & 9.90$\pm$0.38\\
\dataMNIST(GEN) & \costfail{-7.44$\pm$0.10} & -0.08$\pm$0.09 & -0.10$\pm$0.12 & -0.08$\pm$0.10 & \costfail{3.45$\pm$0.09} & 9.90$\pm$0.38 & 9.85$\pm$0.41 & 9.92$\pm$0.34\\
\bottomrule
\end{tabular}
}
\end{small}
\end{center}
\vskip -0.1in
\end{table}
}
\TableAppendixCTSCostFull



\end{document}